%% file: main.tex
\documentclass{article}
\usepackage{float}
\usepackage[eandd,final]{neurips_2026}
\usepackage[american]{babel}
\usepackage[utf8]{inputenc}
\usepackage[T1]{fontenc}
\usepackage{url}
\usepackage{fancyvrb}
\usepackage{fvextra}
\usepackage{booktabs}
\usepackage{amsfonts}
\usepackage{nicefrac}
\usepackage{microtype}
\usepackage{xcolor}
\usepackage{graphicx}
\usepackage{amsmath}
\usepackage{amssymb}
\usepackage{mathtools}
\usepackage{amsthm}
\usepackage{thmtools}
\usepackage{csquotes}
\usepackage{hyperref}
\usepackage{fontawesome5}
\usepackage{dsfont}
\usepackage{cleveref}
\usepackage[most]{tcolorbox}
\usepackage{subcaption}
\usepackage{siunitx}
\usepackage{xurl}
\usepackage{multirow}

\DeclareSIUnit{\billion}{B}
\DeclareSIUnit{\million}{M}
\DeclareSIUnit{\kilo}{k}
\DeclareSIUnit{\token}{tokens}
\DeclareSIUnit{\dollar}{USD}
\DeclareSIUnit{\sample}{sample}
\DeclareSIUnit{\prompt}{prompt}
\DeclareSIUnit{\flop}{FLOP}

\crefname{appendix}{Appendix}{Appendices}
\Crefname{appendix}{Appendix}{Appendices}
\crefname{subappendix}{Appendix}{Appendices}
\Crefname{subappendix}{Appendix}{Appendices}

\tcbuselibrary{listings,skins,breakable}
\usetikzlibrary{positioning,arrows.meta,shapes.geometric,fit,backgrounds,calc,decorations.pathreplacing,shadows.blur}

\graphicspath{{img/}}

\theoremstyle{remark}

\newcommand{\R}{\mathbb{R}}
\DeclareMathOperator*{\E}{\mathbb{E}}
\DeclareMathOperator{\Var}{Var}

\newcommand{\RL}{\mathrm{RL}}
\newcommand{\PT}{\mathrm{PT}}
\newcommand{\SFT}{\mathrm{SFT}}
\newcommand{\EvoLM}{\operatorname{EvoLM}}
\newcommand{\OLMES}{\operatorname{OLMES}}
\newcommand{\auto}{\operatorname{auto}}
\newcommand{\aug}{\operatorname{aug}}
\newcommand{\PolyPythia}{\operatorname{PolyPythia}}

\DeclareMathOperator{\argmin}{\operatorname{arg\,min}}
\DeclareMathOperator{\argmax}{\operatorname{arg\,max}}

\theoremstyle{plain}
\newtheorem{theorem}{Theorem}
\newtheorem{proposition}[theorem]{Proposition}
\theoremstyle{definition}

\title{Welfare, Improvability, and Variance: \\A Principal--Agent Approach \\to Optimal Benchmark Item Aggregation}

\author{%
  Andreas Haupt\thanks{Equal contribution.} \\
  Department of Economics \& Computer Science \\
  Stanford University \\
  \texttt{h4upt@stanford.edu} \\
  \And
  Justin Hartenstein\footnotemark[1] \\
Institute for Computational and Mathematical Engineering \\
Stanford University \\
  \texttt{justinha@stanford.edu} \\
  \And
  Anka Reuel \\
  Department of Computer Science \\
  Stanford University \\
  \texttt{anka@cs.stanford.edu} \\
  \And
  Mykel J.~Kochenderfer\thanks{Equal senior authorship.} \\
  Department of Aeronautics \& Astronautics \\
  Stanford University \\
  \texttt{mykel@stanford.edu} \\
  \And
  Sanmi Koyejo\footnotemark[2] \\
  Department of Computer Science \\
  Stanford University \\
  \texttt{sanmi@cs.stanford.edu} \\
}

\begin{document}

\maketitle

\begin{abstract}
    AI benchmarks have well-documented limitations, with prior work examining contamination, saturation, and construct underspecification. Aggregation has received far less attention: benchmarks are typically summarized by uniformly averaging item-level scores, implicitly treating every test item as equally valuable. We model benchmarking as a multitask principal–agent game and show that the welfare loss from a benchmark is determined jointly by three item-level primitives: alignment with normative welfare priorities, marginal improvability, and performance variance. We translate the theory into an audit framework that ranks items along each of these three axes, and apply it to OLMES items using WORKBank for welfare, the EvoLM 4B suite for improvability, and the PolyPythias 410M panel for variance. The framework surfaces items that are Pareto-inferior within OLMES subject to a pro-worker welfare operationalization. All code is available at~\href{https://github.com/stair-lab/principal-agent-benchmarks}{\faGithub}.
\end{abstract}

\section{Introduction}

AI benchmarks have well-documented limitations, with prior work examining contamination \citep{xu2024benchmark, balloccu2024leak}, saturation \citep{bowman2021what, kiela2021dynabench, ott2022mapping}, and construct underspecification \citep{raji2021ai, salaudeen2025measurement}. Aggregation has received far less attention: benchmarks are typically summarized by uniformly averaging item-level scores \citep{liang2023holistic,hendrycks2021measuring}, implicitly treating every test item as equally valuable. The design choices embedded in that compression---which items to include, and how to weight them---therefore determine which capabilities labs invest in next \citep{perlitz2024efficient, polo2024tinybenchmarks, kipnis2024metabench}.

A growing literature asks what makes a good benchmark \emph{item}, with two largely disjoint streams. The first applies item response theory and information-theoretic informativeness criteria, often paired with construct-validity audits and item-error analysis to screen for questions that genuinely measure a coherent latent ability \citep{liang2023holistic,kipnis2024metabench,truong2025bugs,salaudeen2025measurement}. A second stream documents that benchmarks are not neutral measuring instruments but incentive systems. Selective disclosure, contamination, sandbagging, and prompt-format gaming by labs systematically distort published rankings \citep{singh2025leaderboardillusion,vanderweij2024sandbagging,sainz2023nlp,mizrahi2024state}. The first literature improves benchmarks as measurement instruments; the second shows that benchmarks also function as incentive systems. What remains missing is an item-level account of benchmark quality that conditions on the strategic response induced by a benchmark.

This paper assesses benchmark items taking into account lab incentives, grounded in a game-theoretic model. It solves a game-theoretic model of \emph{optimal benchmark aggregation}. A benchmark designer commits to an item aggregation rule, a lab best-responds with costly and uncertain actions to improve their model---for example, additional pretraining, supervised finetuning, or reinforcement-learning-based finetuning. The model we propose is an adaptation of the multitask principal--agent framework of \citet{holmstrom1991multitask} to benchmarking. Using the framework, we show that the welfare-maximizing aggregation rule takes a closed form and is a combination of three item-level primitives: welfare alignment, improvability, and performance variance.

This theory suggests an auditing schema for benchmark items. First, an item must load on capabilities that are aligned with welfare-relevant concepts, so that effort spent improving it actually produces social value \citep{strathern1997improving}. Second, the dimension must be improvable, since items along intractable axes generate effort-cost without movement. Third, it must have reliable performance in the sense that gains in true capability translate into measurable score gains, since otherwise risk-averse labs may shy away from risks of underperforming models \citep{hu2024reuters}.

We operationalize the optimal benchmark aggregation model. Welfare alignment is estimated from worker preferences over AI-affected tasks in WORKBank \citep{shao2025futureworkaiagents}, mediated through O*NET Generalized Work Activities \citep{onet2026}. Improvability is estimated from per-item benchmark outcomes on the EvoLM 4B model suite \citep{qi2026evolm} and cost estimates for investments from public sources. Variance is estimated from across-seed variance on the PolyPythias 410M pretraining panel \citep{wal2025polypythias}.

We apply our analysis to the Open Language Model Evaluation Standard (OLMES) \citep{gu2025olmes} item set, which contains, among others, the AI2 Reasoning Challenge-Challenge (ARC-Challenge) \citep{clark2018arc}, HellaSwag \citep{zellers2019hellaswag}, Massive Multitask Language Understanding (MMLU) \citep{hendrycks2021measuring}, and Physical Interaction Question Answering (PIQA) \citep{bisk2020piqa}. Our analysis yields four findings. First, OLMES items very infrequently load on GWAs that workers would like to see improved. Second, math benchmarks (\textsc{MathQA}, \textsc{AGIEval-AQuA}, \textsc{MMLU-Math}) are welfare-dominated, none of them are on the worker welfare Pareto frontier subject to our operationalization. Third, improvability is nonuniform, and highest via supervised learning for half of items, reinforcement learning for a third, and pretraining for only $14\%$---this suggests that uniform aggregation would steer labs toward SFT-tractable items at the expense of pretraining-tractable ones, and that incentive-aware benchmarks would mitigate this distortion. Finally, adversarially constructed completion benchmarks (HellaSwag, PIQA, ARC-Challenge) are of meaningfully higher variance than other items in this panel \citep{zellers2019hellaswag, bisk2020piqa, clark2018arc, sakaguchi2021winogrande}.

\subsection*{Related Work}
\label{sec:related-work}

Two literatures bear on the question of how benchmark items should be weighted. The first asks whether items \emph{measure} what we claim they measure; the second asks how labs \emph{respond} to being measured. Each is mature on its own terms, but neither speaks to the design problem of choosing item weights when scores themselves induce costly, strategic capability investments by model creators.

The first strand treats the benchmark as a psychometric instrument and interrogates its construct and criterion validity. Item response theory and information-theoretic informativeness criteria have been used to ask whether items load on a coherent latent ability and to prune those that do not, with companion work on per-item scaling behavior \citep{truongitem,schaeffer2026generative} and on the dimensionality of model capability more broadly \citep{ruan2024observational}. Construct-validity audits, label-error analyses, and critiques of benchmark scope document the measurement-error side of the problem---noisy or mislabeled items, ambiguous targets, and ability constructs that do not survive scrutiny \citep{northcutt2021pervasive,gema2024mmlu,truong2025bugs,jacobs2021measurement,raji2021ai,bowman2021what,kiela2021dynabench,zhou2025lost}. This literature studies benchmark items in isolation: an item is \enquote{good} if it is a precise, valid signal of an underlying ability.

The second strand takes the producer's behavior as the object of analysis. It documents the canonical Goodhart/Campbell pathology---once a measure becomes a target, it ceases to be a good measure---in the form of selective disclosure, contamination, sandbagging, and prompt-format optimization that distort published rankings \citep{singh2025leaderboardillusion,donahue2026yrwr,vanderweij2024sandbagging,dehghani2021benchmark,sainz2023nlp,zhou2023contamination,mizrahi2024state,eriksson2025trust,manheim2018categorizing,thomas2022reliance}. The lessons generalize known multitask-distortion findings from manager incentives \citep{baker2002distortion}, teaching-to-the-test \citep{jacob2005rotten}, healthcare pay-for-performance \citep{campbell2007quality}, and policing statistics \citep{eterno2014crime}: a single scalar index, applied to a multidimensional capability, predictably reallocates effort toward what is measured and away from what is valued. This work motivates calls for benchmark governance but stops short of prescribing how the index itself should be constructed.

We close this gap by adapting the multitask principal--agent framework of \citet{holmstrom1991multitask}---in which a designer weighting noisy measures of effort faces an effort-substitution problem when measures vary in distortion and precision---to benchmark design. Our restriction to static, linear aggregation places us within a tradition studied in economics \citep{holmstrom1987aggregation,carroll2015robustness}. The model identifies three primitives that jointly determine welfare loss under any linear aggregator: welfare-misalignment between item and downstream value, cost-asymmetry across items, and measurement noise. We then audit each primitive empirically. Cost is identified from per-item scaling fits on the EvoLM 4B model suite \citep{qi2026evolm}. Noise---the irreducible run-to-run variation that disincentivizes producer investment---is estimated from the PolyPythias 410M multi-seed panel \citep{wal2025polypythias}. Welfare is anchored in worker-elicited preferences from WORKBank \citep{shao2025futureworkaiagents}, mapped to items via human-validated loadings of generalized work activities.

\paragraph{Outline.} The remainder of the paper proceeds as follows. \Cref{sec:model} develops the principal--agent model of benchmark aggregation and derives the closed-form optimal weights $v^\star = (M + r\Sigma)^{-1} M w$. \Cref{sec:construct} maps the model's primitives---welfare, improvability, and noise---to empirically tractable estimands. \Cref{sec:empirical} describes the operationalization on OLMES items using WORKBank, the EvoLM 4B suite, and the PolyPythias 410M panel. \Cref{sec:results} presents four findings on welfare alignment, math-vs.-general dominance, axis-level cost asymmetries, and adversarial-completion noise. \Cref{sec:conclusion} concludes with limitations and directions for future work. Notation is collected in \Cref{app:notation}. Extensions of our theory are in \Cref{app:foa,app:competition}, details on cost estimation are in \Cref{app:costs} and on welfare elicitation in \Cref{app:prompts}. Robustness checks are in \Cref{app:welfare-robustness}. Finally, \Cref{app:additional_noise_results} contains additional results on variance.

\section{Game-Theoretic Analysis of Optimal Benchmark Aggregation}\label{sec:model}
We present a game-theoretic model of benchmark aggregation. A Benchmark Designer chooses a benchmark \emph{score} that linearly aggregates benchmark items $y_1, y_2, \dots, y_m$, for example items from ARC-AGI \citep{chollet2026arcagi2newchallengefrontier}. A \emph{Lab} best-responds by allocating effort across improvement activities $a_1, a_2, \dots, a_n$, for example pretraining tokens and compute, supervised finetuning prompt-completion pairs and compute, or reinforcement-based finetuning. Denote $y \in \R^m$ the vector of items and $a \in \R^n$ the effort vector. The following model is simplified to be linear-quadratic for expositional purposes, an assumption we relax in \Cref{app:foa}. We assume that the score is 
\begin{equation*}
y = A a + \varepsilon, \qquad \varepsilon \sim \mathcal{N}(0, \Sigma)
\end{equation*}
with a quadratic cost $c(a) = \frac{1}{2} a^\top C a$, where $C \succ 0 \in \R^{n \times n}$ and $\Sigma \succ 0 \in \R^{m \times m}$ are positive definite.

The Designer publishes a linear\footnote{The restriction to linear aggregation and a one-shot game can be partly relaxed. \citet{holmstrom1987aggregation} show that when an agent chooses effort continuously over time and the Designer observes a vector of cumulative signals of the Lab's model, a linear contract in the terminal signal is optimal among all history-dependent contracts within a CARA utility Gaussian model. Linear contracts are also robust if the primitives $C$,$A$, and $\Sigma$ are not known \citep{carroll2015robustness}.} benchmark score $s(y) \coloneqq v^\top y$, with aggregation weights $v \in \R^m$. The Lab has constant absolute risk aversion (CARA) preferences \citep{pratt1964risk} with coefficient $r > 0$, $u(x) = -e^{-rx}$, which gives rise to utility from effort that has a mean-variance form, as we will see below, and which is a common choice for scale-invariant decisionmaking under risk. Risk aversion expresses concerns of the lab that effort may not pay off in the eventual result, as for example in unpredictable training runs \citep{hu2024reuters}.

\paragraph{Lab Effort Choice.} As a feature of CARA and Gaussian noise, the Lab's optimization problem is equivalent to maximizing its mean score minus a risk premium given by the score variance \citep{pratt1964risk}:
\begin{align*}
\E[s(y)\mid a] - c(a) - \frac{r}{2}\Var(s(y)\mid a)
= v^\top A a - \frac{1}{2} a^\top C a - \frac{r}{2} v^\top \Sigma v.
\end{align*}
The first-order condition gives $A^\top v - C a = 0$ and hence a Lab effort of $a^\star(v) = C^{-1} A^\top v$. If we define the \emph{item-level improvability} as $M \coloneqq A C^{-1} A^\top \in \R^{m \times m}$, then the induced performance of a benchmark, taking into account incentives, is $\E[y \mid a^\star(v)] = A a^\star(v) = M v,$ and the induced effort cost is $ \frac{1}{2} (a^\star(v))^\top C a^\star(v) = \frac{1}{2} v^\top M v.$

\paragraph{Aggregation Problem.} Society values benchmark performance $y$ according to welfare weights $w \in \R^m$ and is risk-neutral.\footnote{We focus on the impacts of training investments that fail to result in improvements disincentivizing Lab investment. A model where society is itself CARA risk-averse leads to an identical rule.} We take the Designer to choose a benchmark weighting $v$ to maximize total surplus: social value of induced model improvements minus real effort costs and the compensation-equivalent risk borne by the Lab. Equivalently, the designer internalizes the fact that noisy benchmarks require higher expected returns to induce the same investment from a risk-averse lab. The designer objective is:
\begin{align*}
W(v)
&\coloneqq
\E\left[w^\top y \mid a^\star(v)\right] - c(a^\star(v)) - \frac{r}{2} v^\top \Sigma v =
w^\top M v - \frac{1}{2} v^\top M v - \frac{r}{2} v^\top \Sigma v.
\end{align*}
Hence, the reduced problem is $\max_{v \in \R^m}  w^\top M v - \frac{1}{2} v^\top M v - \frac{r}{2} v^\top \Sigma v$, which is optimized by an optimizer satisfying $(M + r \Sigma) v = M w$, yielding an optimal aggregator
\begin{equation}
\boxed{v^\star = (M + r \Sigma)^{-1} M w.} \label{thm:optimal_v}
\end{equation}
The matrix $M$ summarizes how costly it is to improve tasks jointly: $M_{jk}$ is large when there exist improvement activities $i \in \{1, 2, \dots, n\}$ that raise both tasks $j$ and $k$.
\subsection{Interpreting the Optimal Weights.}
The scalar version of \eqref{thm:optimal_v} shows the features of the optimal benchmark aggregation. If $m = n =1$ and the only entries of $\Sigma$ and $M$ are $s$ and $x$ respectively, it is
\begin{equation}
v^\star = \frac{xw}{x + rs} .\label{eq:scalar-formula}
\end{equation}
This means that the welfare $w$ is scaled by a shrinkage factor governed by an improvability-to-noise ratio $\nicefrac{x}{rs}$: when the activity is highly productive and effort cheap (large $x$), and when measurement is precise and the Lab is risk-tolerant (small $rs$), the Designer transmits welfare directly, $v^\star \approx w$. In the matrix model, the shrinkage factor $\nicefrac{x}{x+rs}$ becomes the matrix operator $(M + r\Sigma)^{-1} M$.

We can formalize \eqref{eq:scalar-formula} in the matrix model through a change of variables if $M$ and $\Sigma$ share an orthonormal eigenbasis, $M = Q \operatorname{diag}(\lambda_1,\dots,\lambda_m) Q^\top$, $\Sigma = Q \operatorname{diag}(\sigma_1,\dots,\sigma_m) Q^\top$, the optimal equation decomposes into independent equations for what we call \emph{concepts} \citep{truong2025bugs}. In the concept coordinates $\hat v^\star = Q^\top v$ and $\hat w = Q^\top w$, the optimal weights are:
\begin{equation}
\hat v_j^\star = \frac{\hat w_j}{1 + \frac{r\sigma_j}{\lambda_j}}.\label{eq:main}
\end{equation}
Equation \eqref{eq:main} means that (i) higher welfare relevance (larger $\hat w_j$), (ii) lower improvement cost (larger $\lambda_j$), and (iii) higher measurement precision (smaller $\sigma_j$) all increase optimal aggregation weight.

\section{Operationalization of Primitives}\label{sec:construct}

The primitives $A$, $C$, $\Sigma$, and $w$ in \Cref{sec:model} are stylized objects, but relate to existing literatures in item-level scaling laws, stochasticity of training run results, and worker preferences on AI development.

\paragraph{Welfare.} The welfare vector $w$ is a normative object. In this paper, we adopt a \emph{pro-worker} stance \citep{acemoglu2023can}, which responds to the economic relevance of AI for work and widespread concerns regarding the impacts of AI on the labor market \citep{brynjolfsson2025canaries}. We do not claim that worker preferences exhaust social welfare. They omit consumers, firms, students, scientific users, safety externalities, and long-run innovation effects. Our narrower goal is to show how a benchmark audit changes under one transparent, in this case labor-centered, welfare criterion. 

There are at least two ways of how AI capabilities can translate into work. AI can augment---help humans do a task---or automate a task. There are many calls for human augmentation in tasks \citep{haupt2025ai,brynjolfsson2022turing}. At the same time there are tasks that workers prefer would be automated, as they are tedious, repetitive, or otherwise unrewarding to perform \citep{shao2025futureworkaiagents}. We will capture this difference by scoring items by how much they load on activities that humans prefer to be automated or augmented. 

\paragraph{Improvability.} Improvability, as captured by $A$ and $C$ in the model, is about how effort, for example pretraining tokens, supervised finetuning prompt-completion pairs, or reinforcement learning rollouts, translate into performance. It is intimately connected to (per-item) scaling laws \citep{hoffmann2022chinchilla,truongitem}. One object we report is the cost of improvement $c_{ik} = \nicefrac{\kappa_k}{(\partial y_i / \partial a_k)}$, the monetary cost of raising performance on an item by one unit through activity $k$, where $\kappa_k$ is the marginal cost of the activity, such as a price per prompt-completion pair.

\paragraph{Noise.} The covariance $\Sigma$ captures any result from variation from the choice of effort to a final outcome. Noise arises both in evaluation---sampled prompts, stochastic decoding, judge variation---and training variation, in particular the consequence of the random seed, which captures random data order, initialization, and stochastic gradients.

The empirical audit does not recover the full structural objects $A, C$ and $\Sigma$ and the normative object $w$. Instead, it estimates item-level proxies for the three forces identified by the model: welfare alignment,  improvability, and measurement noise. These proxies are sufficient for identifying items that are weak along all three primitives.

\section{Empirical Approach}\label{sec:empirical}

We instantiate welfare, cost, and noise for tasks in the OLMES suite~\citep{gu2025olmes}, which contains the Massive Multitask Language Understanding benchmark (\textsc{MMLU}; \citealt{hendrycks2021measuring}), the AI2 Reasoning Challenge---Challenge set (\textsc{ARC-C}; \citealt{clark2018arc}), \textsc{MathQA}, \textsc{HellaSwag}~\citep{zellers2019hellaswag}, \textsc{CommonsenseQA}~\citep{talmor2019commonsenseqa}, the Physical Interaction Question Answering benchmark (\textsc{PIQA}; \citealt{bisk2020piqa}), and the \textsc{AQuA} subset of \textsc{AGIEval}. We instantiate concepts as automation and augmentation of generalized work activities \citep{shao2025futureworkaiagents}, use a human-validated language-model-as-a-judge for concept loading, and take worker preferences over augmentation and automation from \citet{shao2025futureworkaiagents} as two separate concept welfare vectors. For cost we study the sample complexity of improving item-level score by a fixed amount on the EvoLM 4B model suite~\citep{qi2026evolm}, whose models vary in pretraining tokens, supervised finetuning samples, and reinforcement-learning rollouts. For noise, we evaluate item-performance variation across PolyPythias 410M multi-seed checkpoints~\citep{wal2025polypythias}. The following subsections detail our methodology.

\subsection{Worker Welfare}\label{sec:setup-welfare}

To proxy $w$, we use worker preferences over the automation and augmentation of work tasks, collected by \citet{shao2025futureworkaiagents} from \num{1500} workers. For each task $t$, workers report how much they would like AI to automate the task, $A(t) \in [1,5]$, and how much human involvement they would like to retain, $H(t) \in [1,5]$, ranging from wanting AI to carry out the task entirely on its own (\num{1}) to wanting human involvement to remain essential (\num{5}). We aggregate tasks based on a hierarchical taxonomy of work maintained by the U.S. Bureau of Labor Statistics (Occupational Information Network, \citep{onet2026}). We use the highest level of aggregation, generalized work activities (GWAs), and consider the 27 cognitive GWAs due to their relevance to the benchmark items we consider. Examples from this category of GWAs are \emph{Interpreting the Meaning of Information for Others} and \emph{Scheduling Work and Activities}. We drop three of the GWAs containing less than 5 tasks rated in \citep{shao2025futureworkaiagents}. For each GWA $g$, we derive an \emph{automation share} $\auto(g)$ as the fraction of tasks within GWA $g$ on which many workers express strong automation desire. Concretely, we compute it as $\tfrac{1}{|g|}\sum_{t\in g}\mathbf{1}\{A_w(t)\geq 3.5\}$, where the sum is over tasks $t$ that belong to work activity $g$. For each benchmark item, we label the five GWAs the item most strongly loads on. We use a human-validated LLM-as-judge protocol \citep{li-etal-2025-generation} to scale this elicitation. Prompts and human evaluation are provided in \Cref{app:prompts}.

We compute a welfare weight based on the loaded GWAs 
\[
w^{\auto}_i = \sum_{k=1}^5 \lambda_k \mathrm{auto}(g_{ik}).
\]
Here, $g_{ik}$ is the GWA that is loaded $k$th most for item $i$, and $\lambda_k = \frac{1}{k}$ gives a larger weight to more highly ranked GWAs. We compute the augmentation share $w^{\aug}_i$ analogously based on $H(t)$, and normalize both scores to $[1,5]$.

\subsection{Improvability}\label{sec:setup-cost}

To estimate improvability, we use the EvoLM model suite \citep{qi2026evolm}, a family of \num{110} model checkpoints of transformer architectures with 0.5, 1, and 4 billion parameters that systematically vary pretraining tokens, continual pretraining composition, supervised fine-tuning, and reinforcement learning relative to a fixed anchor configuration. We use the 4 billion parameter architecture and one anchor checkpoint that uses \qty{160}{\billion\token} for pretraining, \qty{8}{\billion\token} FineWeb-Edu and \qty{42}{\billion\token} on FineMath, \num{100000} supervised finetuning prompt-completion pairs trained for one epoch, and no reinforcement learning. Continual pretraining is constant, and models vary across their pretraining tokens (3 checkpoints), supervised finetuning prompt-completion pairs (9 checkpoints), and reinforcement learning rollouts (9 checkpoints). The checkpoints increase only in one parameter each; the full Cartesian product is not provided in EvoLM. We therefore will refer to \emph{axes} $k \in \{\PT,\SFT,\RL\}$ as the parts of EvoLM that vary pretraining, supervised finetuning, and reinforcement learning, respectively.

We evaluate OLMES items on all of the checkpoints. As all of OLMES elicits single-token responses, we can evaluate the probability of correctness $p_{im}$ of a checkpoint $m \in \EvoLM$ on an item $i \in \OLMES$ without sampling. To correct for difficulty of items, we fit per-item ordinary least squares regressions
\[
p_{im}= \alpha_{ik} + \beta_{ik} D_{mk} + \varepsilon_{im}
\]
across the available checkpoints $m \in k$ along each axis $k \in \{\PT,\SFT,\RL\}$, where $D_{mk}$ is the respective effort of $m$ within axis $k$, for example, $D_{mk} = \qty{100000}{\sample}$ within axis $k = \SFT$. The different investments become commensurable by translating them into monetary units. We derive the costs of pretraining tokens, supervised-finetuning completions, and reinforcement learning rollouts in \Cref{app:costs} at $\kappa_{\PT}^{4\mathrm{B}} \approx \qty{8.2e-8}{\dollar\per\token}$, $\kappa_{\SFT} \approx \qty{2e-4}{\dollar\per\sample}$, $\kappa_{\RL} \approx \qty{5e-4}{\dollar\per\prompt}$. We compute the marginal improvability in monetary units as $c_{k} = \nicefrac{\kappa_k}{\beta_{k}}$. The minimum cost $\argmin_{k} c_k$ is the cheapest improving intervention and $\min_k c_k$ its cost. We exclude $i \in \OLMES$ where $p_i \notin [0.05, 0.95]$ or $\hat\beta_{ik} < 0$, leading to a total exclusion of \num{2917} items (\qty{16.4}{\percent}).

\subsection{Variance}\label{sec:setup-noise}

We estimate noise using the PolyPythias panel \citep{wal2025polypythias}, which provides multiple independent realizations of the same pretraining run across seeds at sizes up to \qty{410}{\million} parameters. To our knowledge, this is the only public model panel with multi-seed pretraining for multiple checkpoints. For each PolyPythias-410M seed $s = 1, 2, \dots, S$ and checkpoint $m \in \PolyPythia$ on a log-spaced trajectory, we compute the per-item probability $p_{ims}$ as in \Cref{sec:setup-cost}. The per-item noise estimate is the pooled within-checkpoint across-seed empirical variance,
\[
\hat\sigma_i^2 = \frac{1}{N - \lvert \PolyPythia \rvert} \sum_{m \in \PolyPythia} \sum_{s=1}^{S}
\bigl(p_{ims} - \bar p_{im}\bigr)^2,
\]
with $\bar p_{im}$ the across-seed cluster mean and $N = S \lvert \PolyPythia \rvert$. Pooling across checkpoints assumes that across-seed variance is approximately constant across pretraining stages. We use $S = 8$ seeds and $K = 10$ checkpoints and exclude items with $p_{im} \notin [\num{0.10}, \num{0.90}]$ or $\hat\sigma_i^2 < \num{1e-6}$ for numerical reasons. This leads to a total exclusion of \num{11568} of \num{31774} items (\qty{36.4}{\%}).

\section{Results} \label{sec:results}
The methods described in \Cref{sec:empirical} produce, for each OLMES item, two welfare (automation and augmentation), one cost, and one noise score. We provide four findings from them.

\subsection*{Finding 1: Few OLMES Items are Aligned with Worker Welfare}

\begin{figure}[ht]
  \centering
  \includegraphics[width=\linewidth]{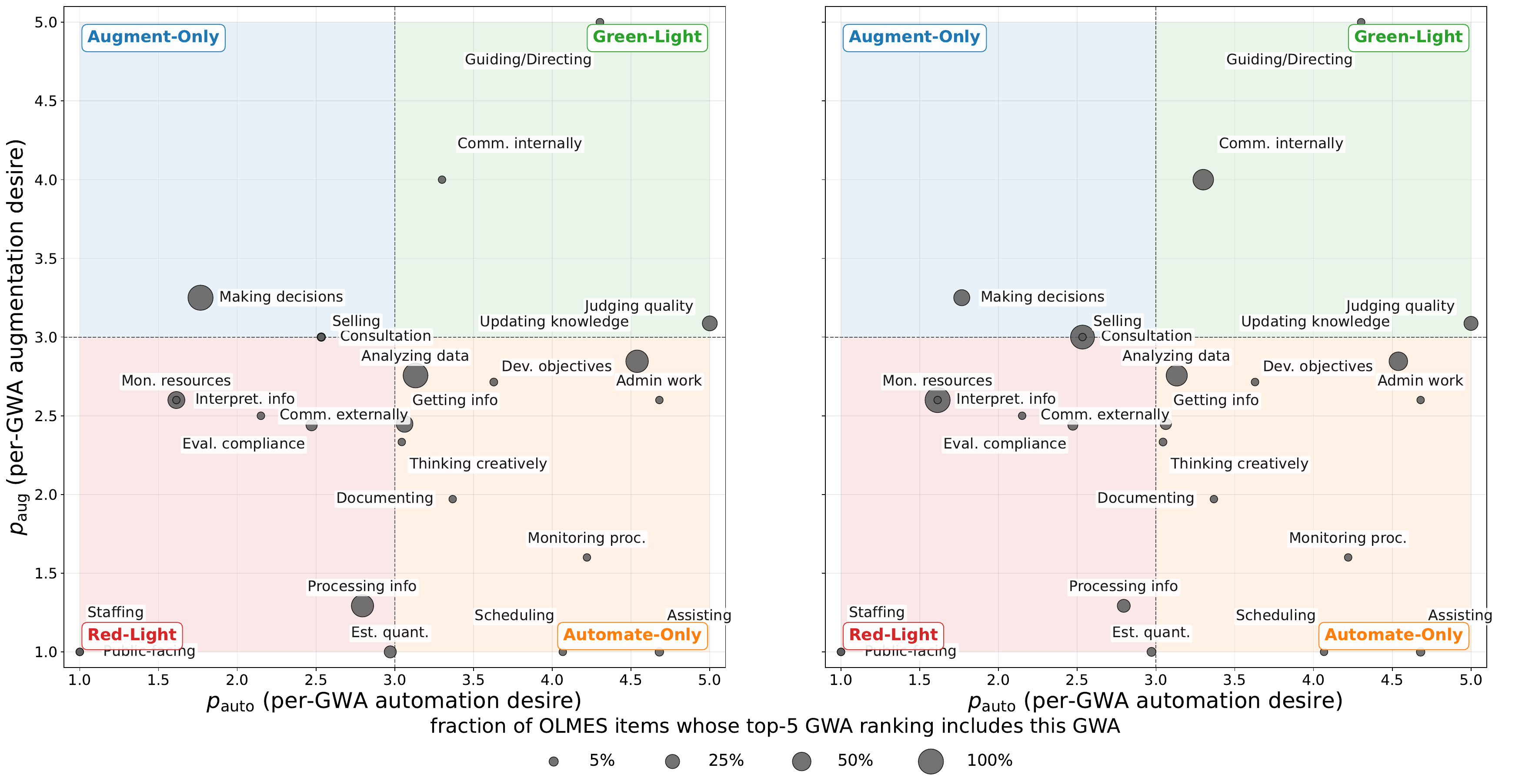}
  \caption{GWA loadings in the WORKBank welfare landscape, under automation-framed (left) and augmentation-framed (right) welfare rankings. Each point is one of 24  GWAs at its per-GWA welfare score; size encodes the fraction of OLMES items whose top-5 ranking includes that GWA. The GWAs in the green-light quadrant (top right), where workers desire both automation and augmentation, receive far fewer loadings than other quadrants.}
  \label{fig:gwa-quadrant}
\end{figure}

Adapting \citep{shao2025futureworkaiagents}, we call \emph{green-light} those work activities where workers have both high automation and augmentation desire. This does not imply that workers simultaneously prefer full automation and human-led augmentation; rather, it indicates that workers view AI involvement favorably under either deployment mode. We map the 24 GWAs we use onto the WORKBank welfare landscape in \Cref{fig:gwa-quadrant}. Then, for each GWA we calculate the share of OLMES items that include the GWA in the item's top-5 ranking. Most items load heavily on the information-processing GWAs (\emph{Getting Information}, \emph{Analyzing Data}, \emph{Processing Information}, and \emph{Making Decisions / Solving Problems}), but only $6\%$ of automation loading mass falls in the green-light quadrant, where workers want model improvement under either deployment mode. Under both framings, $27$--$29\%$ of loading mass falls in the red-light quadrant---high-value activities where workers want little AI involvement or none at all.

\subsection*{Finding 2: Under Pro-Worker Welfare, General-Knowledge Dominate Math Benchmarks}

\begin{figure}[ht]
    \centering
    \includegraphics[width=1\linewidth]{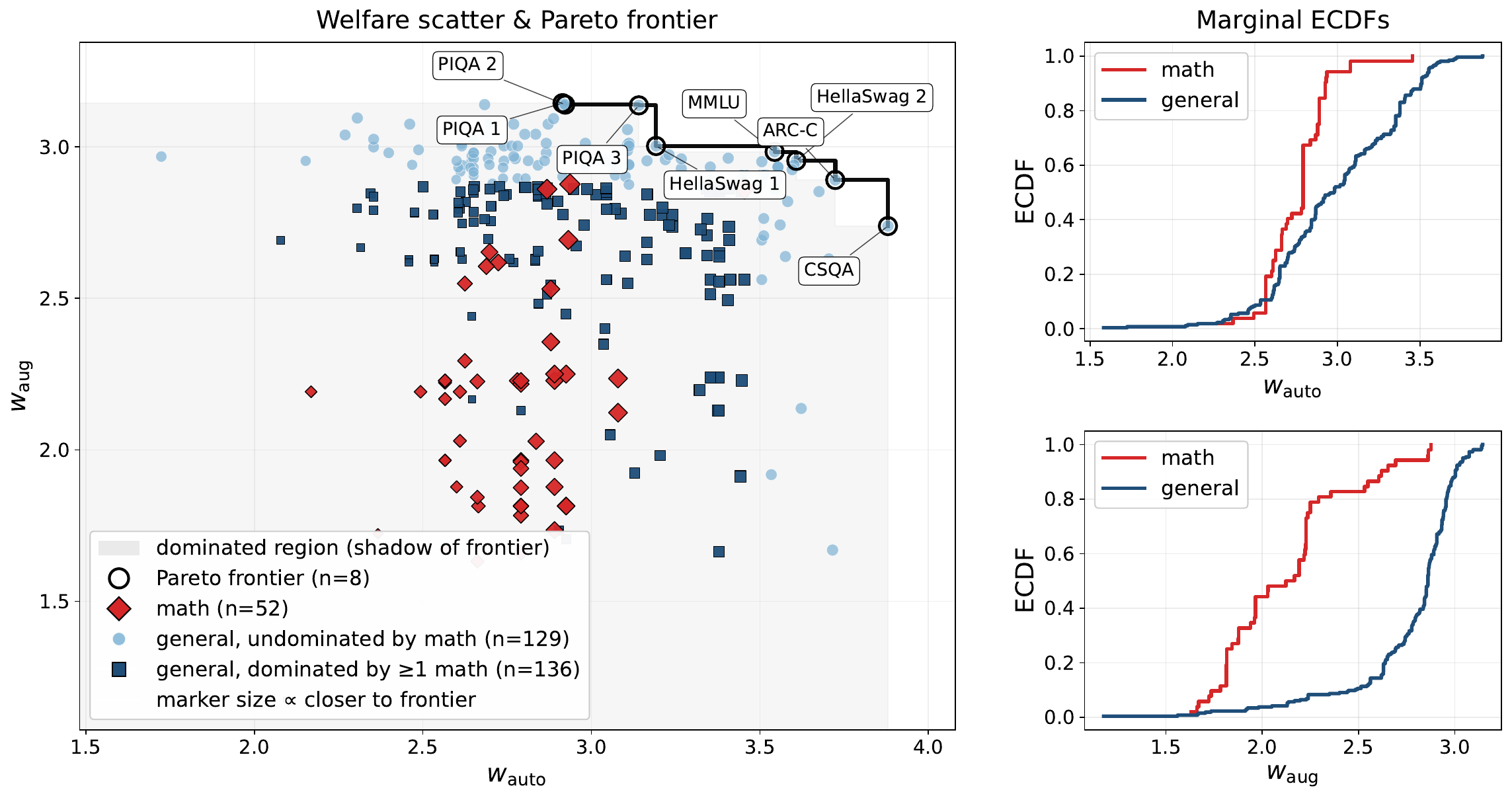}
    \caption{Under our pro-worker welfare operationalization, general-knowledge benchmark items dominate math benchmark items. Per-item welfare scores in $(w^{\auto}, w^{\aug})$ space (higher = higher AI desire; both axes in $[1,5]$). \emph{Left:} all $n=317$ items; the global Pareto frontier ($n=8$) contains general-knowledge benchmarks only, and \qty{100}{\%} of math items ($n=52$) lie in its dominated shadow. \emph{Right:} marginal empirical cumulative distribution functions show general items first-order stochastically dominating math on both axes.}
    \label{fig:welfare_pareto}
\end{figure}

Every item stemming from one of the math benchmarks (\textsc{MathQA}, \textsc{AGIEval}, \textsc{MMLU}-Math) is Pareto-dominated by at least one item from the general-knowledge benchmarks within our panel, and the welfare Pareto frontier contains zero math items (\Cref{fig:welfare_pareto}, panel A). The empirical CDF dominance over the observed item set generalizes to first-order stochastic dominance on both welfare axes (panel B; cluster-bootstrap $p < \num{e-3}$ (augmentation), $p = 0.044$ (automation)): math is worse not just on average but at essentially every quantile threshold. This is consistent with math benchmarks loading on GWAs workers value less on both framings (processing information, solving problems), item-by-item, while general-knowledge benchmarks load on broader capabilities (using information, decision-making,  communication) that rank high on both.

\subsection*{Finding 3: SFT-Improvable Items Are Cheapest Under Our Dollar-Cost Imputation}

Items vary widely in improvability (\Cref{tab:cost-quant-4B}), and the cheapest axis differs across items:  Supervised finetuning is the cheapest intervention for half of all items (\qty{50.3}{\percent}), reinforcement learning for a third (\qty{35.4}{\percent}), and pretraining for only \qty{14.3}{\percent}. The median cost for SFT-cheapest items is \qty{8500}{\dollar} (interquartile range \qtyrange{4.5}{20}{\kilo\dollar}), for RL-cheapest items \qty{104}{\kilo\dollar} (interquartile range \qtyrange{63}{196}{\kilo\dollar}), for PT-cheapest items \qty{250}{\kilo\dollar} (interquartile range \qtyrange{135}{579}{\kilo\dollar}). PT-cheapest items therefore cost a median \num{29} times more than SFT-cheapest items and \num{2.4} times more than RL-cheapest items (both pairwise Mann--Whitney $p < \num{e-200}$).

This is consistent with uniform aggregation steering resources toward SFT-tractable items at the expense of PT-tractable ones. Under uniform weights, the Lab's optimal effort allocation (per Section~\ref{sec:model}) concentrates on the items with the cheapest improving axis---predominantly SFT-tractable items in our panel.

\begin{table}[t]
\centering
\footnotesize
\caption{Sample items from three quantiles according to improvability. Rank CI is at \qtyrange{2.5}{97.5}{\percent} based on $\num{1000}$ residual bootstrap replicates of the per-item ordinary least squares. Axis denotes the cheapest intervention. Cost is the EvoLM cost required for a \qty{10}{\%} improvement.}
\label{tab:cost-quant-4B}
\centering
\begin{tabular}{@{}rllp{0.34\linewidth}lr@{}}
\toprule
{Rank} & {Rank CI} & {Benchmark} & {Topic} & {Axis} & {Cost} \\
\midrule
\multicolumn{6}{l}{\emph{10\% quantile}} \\
\midrule
 1489 & [194, 13373]   & \textsc{HellaSwag}                & performing tricks with trained dogs               & SFT & \$385 \\
 1490 & [1270, 1790]   & \textsc{MMLU}/public relations    & public relations communication theories           & SFT & \$385 \\
 1491 & [1159, 1973]   & \textsc{MMLU}/high school physics & calculating buoyancy and floating objects         & SFT & \$385 \\
\midrule
\multicolumn{6}{l}{\emph{50\% quantile}} \\
\midrule
 7452 & [5657, 14646]  & \textsc{MathQA}                   & finding a number from a ratio and LCM             & RL  & \$4.2k \\
 7453 & [6860, 9080]   & \textsc{MMLU}/philosophy          & Plato's theory of knowledge and forms             & RL  & \$4.2k \\
 7454 & [5042, 14498]  & \textsc{ARC-C}                    & independent variables on a graph                  & RL  & \$4.2k \\
\midrule
\multicolumn{6}{l}{\emph{90\% quantile}} \\
\midrule
 13415 & [11484, 14719] & \textsc{CSQA}                    & finding shelter for human survival                & PT  & \$36k \\
 13416 & [10117, 13899] & \textsc{MMLU}/prof.\ accounting  & modifying  report for financial stmts. & RL  & \$37k \\
 13417 & [11820, 14579] & \textsc{MMLU}/human sexuality    & Freudian concepts of human sexuality              & PT  & \$37k \\
\bottomrule
\end{tabular}
\end{table}

\subsection*{Finding 4: Adversarial Completion Benchmarks Are Noisier Targets than Short-Form Multiple-Choice Questions in PolyPythia-410M}

\begin{table}[t]
\centering
\footnotesize
\caption{Sample items from three quantiles according to variance. Rank CI is at \qtyrange{2.5}{97.5}{\percent} based on $\num{1000}$ across-seed bootstrap samples.}
\label{tab:noise-quant-4B}
\setlength{\tabcolsep}{4pt}
\begin{tabular}{@{}rllp{0.38\linewidth}rr@{}}
\toprule
Rank & Rank CI & Benchmark & Topic & $\hat\sigma$ & $\bar p$ \\
\midrule
\multicolumn{6}{l}{\textit{10\% quantile}} \\
\midrule
 2019 & [629, 5999]   & MMLU/macroeconomics & increase in money demand factors           & 0.057 & 0.14 \\
 2020 & [703, 5478]   & MMLU/philosophy     & Aquinas' views on pleasure and operation   & 0.057 & 0.13 \\
 2021 & [404, 6542]   & MMLU/nutrition      & facial flushing, ALDH enzyme deficiency    & 0.057 & 0.16 \\
\midrule
\multicolumn{6}{l}{\textit{50\% quantile}} \\
\midrule
 10100 & [6456, 11376] & MMLU/college medicine & child psychology evaluation signs        & 0.078 & 0.19 \\
 10101 & [1029, 12391] & MMLU/high school math & probability of divisibility by 2 and 3   & 0.078 & 0.12 \\
 10102 & [5387, 11716] & CSQA                  & habitat and environment of crabs         & 0.078 & 0.22 \\
\midrule
\multicolumn{6}{l}{\textit{90\% quantile}} \\
\midrule
 18183 & [17279, 18875] & AGIEval-AQuA & average speed calculation for a plane         & 0.215 & 0.81 \\
 18184 & [17714, 18609] & MathQA       & total matches in knockout tournament          & 0.215 & 0.40 \\
 18185 & [13869, 19161] & HellaSwag    & treatment options for chicken pox scars       & 0.215 & 0.19 \\
\bottomrule
\end{tabular}
\end{table}

The per-item noise estimate $\hat\sigma_i$ spans almost an order of magnitude across the eligible item pool (\Cref{tab:noise-quant-4B}), and correlates with benchmark membership: Within OLMES, HellaSwag, PIQA, and ARC-Challenge are overrepresented in the noisy tail, while MathQA, MMLU, and CommonsenseQA dominate the clean head. The pattern survives controlling for item difficulty via Bernoulli regression on $\hat p_i(1-\hat p_i)$: HellaSwag retains about three times the residual noise of  \textsc{MMLU} (\Cref{tab:noise-decomp} in \Cref{app:additional_noise_results}). One plausible mechanism is the  log-probability gap between the gold answer and its closest distractor, which  appears systematically larger in short-form multiple-choice questions than in adversarially-constructed benchmarks (HellaSwag, PIQA), where distractors were generated to be plausible to weak models, so small seed-induced shifts in token probabilities flip the model's preference (see \Cref{app:additional_noise_results}).

\section{Conclusion} \label{sec:conclusion}

\paragraph{Summary.} This paper develops a principal--agent theory of AI benchmark design and operationalizes it to understand the benefits of different items in the OLMES benchmark suite. We derive a closed-form expression for optimal benchmark aggregation in our theoretical model, $v^\star = (M+r\Sigma)^{-1}Mw$ in which welfare alignment, improvability, and noise jointly determine how heavily each item should count. We translate the three primitives into separate empirical estimands and apply them to OLMES items using WORKBank, the EvoLM 4B suite, and the PolyPythias 410M panel. The audit surfaces item families that are weak along individual primitives and therefore would tend to receive lower weight in incentive-aware aggregation, all else equal.

\paragraph{Limitations.} Several caveats bound the present work. First, the mapping from benchmark items to occupational exposure relies on an LLM-as-judge classification against O*NET GWAs; this inherits known pathologies of LLM raters. Second, our welfare functional $w$ aggregates worker-side preferences only, excluding non-worker stakeholders. Third, the item-level tests underlying our rankings assume approximate independence across items conditional on model and seed; in practice, items within a benchmark share construction pipelines, annotators, and topical clusters, so reported $p$-values should be read as upper bounds on evidential strength. Fourth, the cost panel (EvoLM 4B) and noise panel (PolyPythias 410M) sit several orders of magnitude below frontier scale and come from different model families; the direction of bias from this scale gap is uncertain---scaling could attenuate noise sensitivities (as capabilities saturate easy items) or amplify them (as frontier evaluations concentrate on harder items)---and we cannot adjudicate between these without a frontier-scale multi-seed replication, which is not publicly available. Finally, the model itself rests on linearity, lab monopoly, and a common production technology across labs; \Cref{app:foa} shows that the closed-form aggregation rule survives nonlinear $\mu$ and $c$ as a fixed-point equation evaluated at equilibrium effort, and \Cref{app:competition} shows that two-lab contest competition replaces $r$ with a contest-implied constant but leaves the rule's algebraic shape intact.

\paragraph{Future work.} Three directions extend the framework naturally. The most immediate is to commit to values of $r$ and a welfare framing and report the single canonical $v^\star$, alongside its sensitivity to those choices. A second is theoretical: the model treats only a single benchmark designer, but in practice labs select which benchmarks to submit to and which scores to disclose \citep{singh2025leaderboardillusion}, and the resulting two-sided selection problem---benchmarks compete for submissions while labs select for flattering weights---is unresolved. A third is infrastructural: the per-item welfare, cost, and noise scores compose naturally with EvalEval-style \citep{evaleval2024evalcards} eval cards as a metadata layer that downstream curators, regulators, and benchmark designers can build on.

\section*{Acknowledgements and Disclosures}
AH is supported by Schmidt Sciences. JH is supported by the German Academic Exchange Service
(DAAD). AR is supported by a Stanford Interdisciplinary Graduate Fellowship. SK is partially supported by NSF 2046795, 2205329, and 2504264, the MacArthur Foundation, Coefficient Giving, Schmidt Sciences, the Hasso Plattner Foundation, and Stanford HAI.

\bibliographystyle{plainnat}
\bibliography{refs}

\appendix

\crefalias{section}{appendix}
\crefalias{subsection}{subappendix}

\section{Notation Reference}\label{app:notation}
\Cref{tab:notation} gives an overview of all notation used in the optimal benchmark aggregation problem.
\begin{table}[!ht]
\centering
\caption{Notation used throughout the model.}
\label{tab:notation}
\footnotesize
\begin{tabular}{@{}p{0.08\linewidth}p{0.22\linewidth}p{0.62\linewidth}@{}}
\toprule
Symbol & Space & Meaning \\
\midrule
\multicolumn{3}{@{}l}{\textit{Primitives}} \\
\midrule
$n$ & $\mathbb{N}$ & Number of effort dimensions, e.g., pretraining, SFT \\
$m$ & $\mathbb{N}$ & Number of benchmark items \\
$a$ & $\mathbb{R}^n$ & Lab's effort vector across activities \\
$y$ & $\mathbb{R}^m$ & Vector of benchmark item outcomes \\
$A$ & $\mathbb{R}^{m \times n}$ & Technology matrix mapping effort to expected performance \\
$\varepsilon$ & $\mathbb{R}^m$ & Gaussian noise vector, $\varepsilon \sim \mathcal{N}(0,\Sigma)$ \\
$\Sigma$ & $\mathbb{R}^{m \times m}$, $\succ 0$ & Noise covariance, benchmark sampling and seed variation \\
$C$ & $\mathbb{R}^{n \times n}$, $\succ 0$ & Quadratic effort-cost matrix \\
$c(a)$ & $\mathbb{R}$ & Lab's cost of effort, $c(a) = \frac{1}{2} a^\top C a$ \\
$r$ & $\mathbb{R}_{\ge 0}$ & Lab's CARA risk-aversion coefficient \\
$w$ & $\mathbb{R}^m$ & Society's welfare weights over items \\
\midrule
\multicolumn{3}{@{}l}{\textit{Designer's instrument and Lab's response}} \\
\midrule
$v$ & $\mathbb{R}^m$ & Designer's linear aggregation weights \\
$s(y)$ & $\mathbb{R}$ & Published benchmark score, $s(y) = v^\top y$ \\
$a^\star(v)$ & $\mathbb{R}^n$ & Lab's best-response effort, $a^\star(v) = C^{-1} A^\top v$ \\
$W(v)$ & $\mathbb{R}$ & Total social welfare under weights $v$ \\
$v^\star$ & $\mathbb{R}^m$ & Optimal aggregation weights, $v^\star = (M + r\Sigma)^{-1} M w$ \\
\midrule
\multicolumn{3}{@{}l}{\textit{Reduced-form objects}} \\
\midrule
$M$ & $\mathbb{R}^{m \times m}$, $\succeq 0$ & Item-level cost structure, $M \coloneqq A C^{-1} A^\top$; $M_{jk}$ large when shared activities raise items $j$ and $k$ jointly \\
$Q$ & $\mathbb{R}^{m \times m}$, orthogonal & Concept eigenbasis (shared eigenbasis of $M$ and $\Sigma$) \\
$\lambda_j$ & $\mathbb{R}_{\ge 0}$ & $j$th eigenvalue of $M$ (improvability of concept $j$) \\
$\sigma_j$ & $\mathbb{R}_{> 0}$ & $j$th eigenvalue of $\Sigma$ (measurement noise on concept $j$) \\
$\hat v$ & $\mathbb{R}^m$ & Aggregation weights in concept coordinates, $\hat v = Q^\top v$ \\
$\hat w$ & $\mathbb{R}^m$ & Welfare weights in concept coordinates, $\hat w = Q^\top w$ \\
\bottomrule
\end{tabular}
\end{table}

\section{Beyond Linear-Quadratic Utility}\label{app:foa}

In \Cref{sec:model} we worked with linear performance $\mu(a) = Aa$ and quadratic cost $c(a) = \frac{1}{2} a^\top C a$. We now drop both assumptions, keeping CARA preferences and additive Gaussian noise whose covariance does not depend on effort. The aggregation rule of \Cref{thm:optimal_v} survives as a first-order condition. The constant matrix $M=AC^{-1}A^\top$ is replaced by a local \emph{improvability matrix} $M(v)$, which depends on $v$ through the Lab's equilibrium effort and through the curvature of $\mu$, and the closed form becomes a fixed-point condition that every interior optimum satisfies.

\paragraph{Setup.} Fix a benchmark with $m$ items and $n$ activities. The Lab chooses $a \in \R^n$ and produces $y = \mu(a) + \varepsilon$ with $\varepsilon \sim \mathcal{N}(0, \Sigma)$, $\Sigma\succ0$, where $\mu : \R^n \to \R^m$ is twice continuously differentiable. The Lab's effort cost $c : \R^n \to \R$ is twice continuously differentiable and convex. The Designer publishes a linear score $s(y) = v^\top y$, and the Lab is CARA with risk-aversion coefficient $r \ge 0$. Society values items according to welfare weights $w \in \R^m$.

\paragraph{Lab's problem.} Because the noise is Gaussian and independent of effort, the Lab's certainty equivalent is
\begin{equation*}
v^\top \mu(a) - c(a) - \frac{r}{2} v^\top \Sigma v,
\end{equation*}
and its first-order condition is
\begin{equation}
\nabla \mu(a^\star)^\top v = \nabla c(a^\star). \label{eq:foc-app}
\end{equation}
Define the Jacobian $J(a) \coloneqq \nabla \mu(a) \in \R^{m \times n}$, the cost Hessian $H(a) \coloneqq \nabla^2 c(a) \in \R^{n \times n}$, and
\begin{equation*}
H_{\mathrm{eff}}(a,v) \coloneqq H(a) - \sum_{i=1}^m v_i \nabla^2 \mu_i(a),
\end{equation*}
so that $-H_{\mathrm{eff}}$ is the Hessian of the Lab's objective. The first-order condition is necessary but not sufficient. All the following assumes that the Lab's objective has a unique maximizer $a^\star(v)$, the map $a^\star(\cdot)$ is continuous, and $H_{\mathrm{eff}}(a^\star(v),v)\succ0$.

\paragraph{Improvability.} The implicit function theorem applied to \eqref{eq:foc-app} shows that $a^\star$ is continuously differentiable on $V$, with
\begin{equation}
\frac{\partial a^\star}{\partial v} = H_{\mathrm{eff}}^{-1} J^\top,
\label{eq:comp-stat-app}
\end{equation}
where $J$, $H$, and $\nabla^2\mu_i$ are evaluated at $a^\star(v)$. The induced sensitivity of expected scores to weights is
\begin{equation*}
M(v) \coloneqq \frac{\partial  \E[y]}{\partial v} = J H_{\mathrm{eff}}^{-1} J^\top \succeq 0,
\end{equation*}
which we call the improvability matrix. In the linear-quadratic case $H_{\mathrm{eff}} = C$ and $J = A$, so $M(v) = A C^{-1} A^\top$ recovers the matrix from \Cref{sec:model} and does not depend on $v$. In general $M(v)$ differs from the cost-only matrix $JH^{-1}J^\top$ evaluated at equilibrium effort, through the curvature term $\sum_i v_i\nabla^2\mu_i$; the comparison requires $H$ to be invertible, which holds, for example, whenever $\sum_i v_i\nabla^2\mu_i \succeq 0$, since then $H \succeq H_{\mathrm{eff}} \succ 0$.

\paragraph{Designer's problem.} Total welfare under $v$ is
\begin{equation*}
W(v) = w^\top \mu(a^\star(v)) - c(a^\star(v)) - \frac{r}{2} v^\top \Sigma v.
\end{equation*}
Differentiating in $v$ using \eqref{eq:comp-stat-app} and the Lab's first-order condition \eqref{eq:foc-app},
\begin{equation*}
\nabla W(v) = M(v)(w - v) - r \Sigma v .
\end{equation*}

\begin{proposition}\label{prop:foa-fixed-point}
Every maximizer $v^\star\in V$ of $W$ satisfies
\begin{equation}
\bigl(M(v^\star) + r \Sigma\bigr) v^\star = M(v^\star) w.
\label{eq:vstar-fp}
\end{equation}
If $r>0$, then $M(v^\star)+r\Sigma\succ0$, and \eqref{eq:vstar-fp} is equivalent to
\begin{equation}
v^\star = \bigl(M(v^\star) + r \Sigma\bigr)^{-1} M(v^\star) w.
\label{eq:vstar-app}
\end{equation}
\end{proposition}

\begin{proof}
On $V$, $W$ is continuously differentiable with the gradient above, so a maximizer in the open set $V$ satisfies $\nabla W(v^\star)=0$, which rearranges to \eqref{eq:vstar-fp}. Since $M(v^\star)\succeq0$ and $\Sigma\succ0$, $M(v^\star)+r\Sigma\succ0$ for $r>0$.
\end{proof}

This has the same algebraic shape as \Cref{thm:optimal_v}, with the global $M$ replaced by its local counterpart $M(v^\star)$. In the linear-quadratic case $M(v^\star) = A C^{-1} A^\top$ for all $v$, and \eqref{eq:vstar-app} is an explicit closed form. In general it is a fixed-point condition, and it is necessary but not sufficient. $W$ need not be concave, so the condition can have several solutions, and comparative statics in $r$, $\Sigma$, and $w$ pick up feedback through $\partial M/\partial v$. In practice one can maximize $W$ directly, since evaluating $W(v)$ requires only the Lab's best response. The concept interpretation carries through at $v^\star$ whenever $M(v^\star)$ and $\Sigma$ commute, that is, share an orthogonal eigenbasis.

\section{Lab Competition}\label{app:competition}
The baseline model treats a single Lab with payoff linear in the score. We now introduce both \emph{competition}---two symmetric Labs racing on the same benchmark---and \emph{ordinal} payoffs, capturing the leaderboard reality that the top model captures disproportionate attention, API traffic, and follow-on contracts. The aggregation rule of \Cref{thm:optimal_v} survives in form, with the risk-aversion coefficient $r$ replaced by an endogenous scalar $\tilde r$: the shadow price of the incentive power that the contest delivers. Unlike $r$, $\tilde r$ can be negative.

Two symmetric Labs $k\in\{1,2\}$ simultaneously choose effort vectors $a^k\in\R^n$ at quadratic cost $\frac{1}{2}(a^k)^\top C a^k$, $C\succ 0$, producing scores $y^k=Aa^k+\varepsilon^k$ with $\varepsilon^k\sim\mathcal{N}(0,\Sigma)$, $\Sigma\succ 0$, i.i.d.\ across Labs. In this extension, we read $\varepsilon^k$ as evaluation noise, so deploying Lab $k$'s model yields value $w^\top A a^k$. The Designer publishes weights $v$ and awards prize $w_H$ to the higher score and $w_L<w_H$ to the loser; Labs are risk-neutral. Win probabilities are invariant to positive rescaling of $v$, so we normalize $v^\top\Sigma v=1$. As in the baseline, $M\coloneqq AC^{-1}A^\top$; $\phi$ and $\Phi$ denote the standard normal density and distribution function.

\paragraph{Lab Equilibrium.} A profile $(a^1,a^2)$ is a (pure-strategy) Nash equilibrium if each Lab's choice maximizes its own expected prize net of cost given the opponent's choice:
\begin{equation*}
a^k\in\argmax_{a\in\R^n}\E\left[w_H\mathds{1}\{v^\top y^k> v^\top y^{-k}\}+w_L\mathds{1}\{v^\top y^k\le v^\top y^{-k}\}\right]-\tfrac{1}{2}a^\top Ca,\quad k=1,2,
\end{equation*}
where $y^k=Aa+\varepsilon^k$ inside the expectation. Lab 1's score advantage $v^\top(y^1-y^2)$ has Gaussian noise with variance $2v^\top\Sigma v=2$, so its win probability is $\Phi(\frac{1}{\sqrt{2}}v^\top A(a^1-a^2))$ and its first-order condition is
\begin{equation*}
\frac{w_H-w_L}{\sqrt{2}}\,\phi\Bigl(\tfrac{1}{\sqrt{2}}v^\top A(a^1-a^2)\Bigr)A^\top v=Ca^1 .
\end{equation*}
Lab 2's condition is the same with $a^1$ and $a^2$ exchanged. Because $\phi$ is even, the two conditions give $Ca^1=Ca^2$, so every pure-strategy equilibrium is symmetric. At a symmetric profile the argument of $\phi$ is zero and $\phi(0)=1/\sqrt{2\pi}$, so
\begin{equation}
a^\star(v) = \beta C^{-1} A^\top v, \qquad \beta\coloneqq\frac{w_H-w_L}{2\sqrt{\pi}}.
\label{eq:contest-foc}
\end{equation}
The first-order condition is necessary but not sufficient: a pure-strategy equilibrium exists only if the noise is sufficiently dispersed relative to the prize spread; see \citet[fn.~2]{lazear1981rank} and \citet[\S 2.2]{konrad2009strategy}. We assume throughout that $(a^\star(v),a^\star(v))$ is an equilibrium at the weights under consideration.

\paragraph{Optimal Aggregation.} Only the winner's model is deployed and prizes are transfers, so welfare is
\begin{equation*}
W(v)=\beta\,w^\top Mv-\beta^2\,v^\top Mv
\end{equation*}
(one unit of expected deployment value, two units of summed effort cost). It helps to work with the effective piece rate $b\coloneqq\beta v$, which plays the role of $v$ in the baseline because $a^\star=C^{-1}A^\top b$. Then $W=2\bigl[(\tfrac{w}{2})^\top Mb-\tfrac{1}{2}b^\top Mb\bigr]$ and the normalization becomes $b^\top\Sigma b=\beta^2$. The prize spread thus fixes the noise-adjusted power of incentives at $\beta^2$, whereas the baseline lets the Designer choose that power and charges the risk premium $\frac{r}{2}b^\top\Sigma b$ for it. Maximizing $W$ subject to $v^\top\Sigma v=1$ with multiplier $\mu$, the stationarity condition $\beta Mw=2\beta^2Mv+2\mu\Sigma v$ gives, with $\tilde r\coloneqq\mu/\beta^2$, $(M+\tilde r\Sigma)v^\star=Mw/(2\beta)$. At the maximizer $M+\tilde r\Sigma\succeq0$, and whenever it is nonsingular,
\begin{equation}
v^\star=\frac{1}{2\beta}\bigl(M+\tilde r\Sigma\bigr)^{-1}Mw,
\qquad\text{equivalently}\qquad
b^\star=\bigl(M+\tilde r\Sigma\bigr)^{-1}M\,\frac{w}{2}.
\label{eq:contest-vstar}
\end{equation}
This is the formula of \Cref{thm:optimal_v} for the value vector $w/2$ (effort is duplicated across Labs, but only one model is deployed), with $\tilde r$ in place of $r$. When $M\succ0$, $\tilde r<0$ exactly when $w^\top\Sigma w<(w_H-w_L)^2/\pi$.

\section{Cost Estimates}
\label{app:costs}

This appendix estimates the marginal cost in \unit{\dollar} for pretraining, supervised finetuning, and reinforcement learning-based finetuning of EvoLM's \qty{4}{\billion} parameter model. All values are estimates for consumer cloud and do not attempt to estimate frontier cloud training.

\paragraph{Cost of a \unit{\flop}.} EvoLM's pretraining and reinforcement learning stages both consume GPU-hours; we anchor both to commodity A100-80GB cloud pricing of \qty{1.5}{\dollar/\hour} \citep[e.g.,][]{jiang2025thunderservehighperformancecostefficientllm} at Nvidia A100 brain float 16-bit (BF16) peak throughput of $\num{3.1e14}$ floating-point operations per second. Assuming a typical \qty{40}{\%} model-FLOPs-utilization for dense training, the cost of a FLOP is
\[
\frac{\SI{1.50}{\dollar/\hour}}{\SI{3.1e14}{\flop/\second} \cdot 0.40 \cdot \SI{3600}{\second/\hour}} \approx \SI{3.4e-18}{\dollar/\flop}
\]
This figure prices both pretraining and reinforcement learning computation. Reinforcement learning rollouts are priced at a lower effective model \unit{\flop} utilization (see below).

\paragraph{Pretraining cost per token.} The Chinchilla rule \citep{hoffmann2022chinchilla} gives training FLOPs per token of $6N$ for a model with $N$ parameters. Combining with the cost per FLOP above,
\begin{align*}
\kappa_{\text{pt}}^{1\text{B}} &= 6 \cdot \num{e9} \cdot \num{3.4e-18} \approx \SI{2.0e-8}{\dollar\per\token} \\
\kappa_{\text{pt}}^{4\text{B}} &= 6 \cdot \num{4e9} \cdot \num{3.4e-18} \approx \SI{8.2e-8}{\dollar\per\token}
\end{align*}
For reference, training EvoLM-4B on its \qty{320}{\billion\token} pretraining budget would cost about \qty{26000}{\dollar} at this rate, consistent with academic-scale \qty{4}{\billion}-model training-cost estimates.

\paragraph{SFT cost per sample.} EvoLM's supervised finetuning data is constructed from a mixture of MetaMathQA \citep{yu2024metamath}, OpenMathInstruct-2 \citep{toshniwal2025openmathinstruct}, and NuminaMath \citep{numina_math_datasets}, all math-targeted instruction datasets generated by frontier LLMs. The largest and most recent of these, OpenMathInstruct-2, was synthesized using Llama-3.1-405B-Instruct. We price the marginal replacement cost of one math problem-solution pair by treating its generation as an API call at frontier quality. At Amazon Bedrock pricing for Llama-3.1-405B of \qty{2.4}{\dollar/\million\token} \cite{aws_bedrock_pricing} a typical math problem-solution pair of 50 input and 200 output tokens costs
\[
\kappa_{\SFT} \approx 50 \cdot \qty{2.40e-6}{\dollar} \cdot 0.75 + 200 \cdot \qty{2.40e-6}{\dollar} \cdot 0.25 \approx \qty{2e-4}{\dollar\per\sample}.
\]
For the \num{500000}-sample supervised finetuning pool used in EvoLM, this implies a replacement cost of about \qty{100}{\dollar}. We use API generation cost as the anchor because it represents what a Lab would pay at the margin to acquire one additional sample at the supervised finetuning  recipe's quality level.

\paragraph{Reinforcement learning cost per prompt.} For verifiable-reward reinforcement learning on math (the EvoLM setup \citep{qi2026evolm}, using Proximal Policy Optimization \citep{schulman2017proximal} with a binary correctness reward), each prompt-rollout-update consists of generating $K$ rollouts of $G$ tokens followed by a gradient update on the resulting $KG$ tokens. With representative settings $K=8$, $G=512$ at the 4B scale:
\begin{align*}
\text{rollout (forward only)} &: K G \cdot 2N \approx \qty{3.3e13}{\flop} \\
\text{gradient update} &: K G \cdot 6N \approx \qty{1.0e14}{\flop}
\end{align*}
Rollouts are autoregressive generation, which typically runs at lower model-FLOPs-utilization than dense training due to memory-bandwidth limits during decoding \citep{pope2023efficiently}. Pricing rollouts at the lower effective MFU and gradient updates at the dense-training MFU,
\[
\kappa_{\RL} \approx \qty{5e-4}{\dollar\per\prompt}.\footnote{We assume \qty{1.50}{\dollar\per\hour} per GPU, A100-class peak throughput of \qty{3.1e14}{\flop\per\second} (BF16), training MFU of $0.40$, and rollout MFU of $0.20$ (achievable with continuous-batched inference, e.g. vLLM \citep{kwon2023efficient}). The settings $K=8$ and $G=512$ are representative for short-CoT math RL; the exact EvoLM hyperparameters are immaterial to this order-of-magnitude estimate.}
\]
Verifier cost is negligible for verifiable-reward math (an answer-check function on the model's output).

\section{Details of GWA Assignment}
\label{app:prompts}

For concept loading, we rely on a language model judge. We validate the protocol against two independent human annotators (denoted A1 and A2), who each ranked the 24 cognitive GWAs by relevance for 28 OLMES items. We then elicit comparable rankings from Claude Opus 4.5 under a 3-example few-shot prompt and report pairwise agreement under both the automation and augmentation framings. To benchmark the LLM judge against the inter-human baseline, we compute nDCG@5 (linear and binary relevance), Precision@4, Recall@4, Jaccard@5, and Rank-Biased Overlap (RBO at $p = 0.7$) for all three ranker pairs. Macro-averages are reported in \Cref{tab:gwa_ranking_auto_aug}.

Two patterns are worth flagging. First, inter-human agreement on OLMES items is moderate (nDCG@5 of \num{0.72} under automation, \num{0.59} under augmentation), which is consistent with augmentation being a less sharply defined construct than automation---a pattern we also see when comparing Claude to either human. Second, Claude's agreement with the two annotators is asymmetric: agreement with A1 exceeds agreement with A2 under both framings, with the gap largest under augmentation (\num{0.70} vs.\ \num{0.60} nDCG@5(linear)). Taking the more conservative of the two LLM-human comparisons of Opus 4.5 vs. A2, Claude tracks human rankings approximately as well as the two humans track each other (0.66 vs. 0.72 under automation; 0.60 vs. 0.59 under augmentation), which we read as suggesting that the LLM-judge protocol recovers a signal of comparable quality to a single human annotator on this item set, while not eliminating annotator-level variation in the construct itself.

\begin{table}[t]
\centering
\small
\caption{Pairwise agreement on GWA rankings across two human annotators (A1, A2) and an LLM judge (Claude with 3-example few-shot), on 28 OLMES items. Higher is better for all metrics.}
\label{tab:gwa_ranking_auto_aug}
\begin{tabular}{@{}lcccccc@{}}
\toprule
{Pair} & {nDCG@5 (lin.)} & {nDCG@5 (bin.)} & {Prec@4} & {Recall@4} & {Jaccard@5} & {RBO} \\
\midrule
\multicolumn{7}{l}{\textit{Automation framing}} \\
A1 $\leftrightarrow$ A2          & 0.723 & 0.743 & 0.705 & 0.668 & 0.548 & 0.555 \\
Opus 4.5 $\leftrightarrow$ A1      & 0.742 & 0.769 & 0.772 & 0.722 & 0.559 & 0.559 \\
Opus 4.5 $\leftrightarrow$ A2      & 0.662 & 0.649 & 0.644 & 0.616 & 0.435 & 0.480 \\
\midrule
\multicolumn{7}{l}{\textit{Augmentation framing}} \\
A1 $\leftrightarrow$ A2          & 0.593 & 0.673 & 0.667 & 0.585 & 0.493 & 0.431 \\
Opus 4.5 $\leftrightarrow$ A1      & 0.700 & 0.747 & 0.749 & 0.676 & 0.592 & 0.530 \\
Opus 4.5 $\leftrightarrow$ A2      & 0.600 & 0.678 & 0.680 & 0.567 & 0.511 & 0.433 \\
\bottomrule
\end{tabular}
\end{table}

We use the following prompts for loading on GWA automation and augmentation.

\begin{tcolorbox}[enhanced,breakable,title=\textbf{Automation-focused Prompt}]
\begin{Verbatim}
An AI system is deployed to handle this task autonomously, with no human \
involvement. Which work activities must the AI exercise to complete this \
item end-to-end?

Work in two steps inside the JSON output below:

STEP 1 — In `sub_steps`, decompose the item from first principles into the \
granular sub-tasks an AI would have to do to arrive at the correct answer. \
Think about the underlying problem, not the surface format (e.g. that it's \
multiple choice).

STEP 2 — In `ranked_gwa_ids`, rank up to 5 GWAs from the list below that \
benefit from being able to do those sub-tasks, ordered from most to least \
central.

Benchmark Item:
{task_text}

---

GWA List:
{gwa_list}

---

Example:

Benchmark Item:
An entrepreneur from State A decided to sell hot sauce to the public, \
labeling it "Best Hot Sauce." A company incorporated in State B and \
headquartered in State C sued the entrepreneur in federal court in State C. \
The complaint sought $50,000 in damages and alleged that the entrepreneur's \
use of the name "Best Hot Sauce" infringed the company's federal trademark. \
The entrepreneur filed an answer denying the allegations, and the parties \
began discovery. Six months later, the entrepreneur moved to dismiss for \
lack of subject-matter jurisdiction. Should the court grant the \
entrepreneur's motion?
  A) No, because the company's claim arises under federal law.
  B) No, because the entrepreneur waived the right to challenge subject-matter jurisdiction by not raising the issue initially by motion or in the answer.
  C) Yes, because although the claim arises under federal law, the amount in controversy is not satisfied.
  D) Yes, because although there is diversity, the amount in controversy is not satisfied.

JSON output:
{{
  "sub_steps": [
    "user pastes the fact pattern and four answer choices, asking which is correct",
    "AI identifies that this is a federal subject-matter jurisdiction question and surfaces the two grounds (federal question vs. diversity)",
    "AI flags the salient facts: federal trademark claim, $50,000 amount, parties from different states",
    "AI checks the live legal rule — federal-question jurisdiction has no amount-in-controversy threshold; SMJ can be raised at any time",
    "user asks whether the timing of the motion matters; AI confirms it does not for SMJ and cites the relevant procedural rule",
    "AI recommends option A and explains why federal-question grounds jurisdiction directly"
  ],
  "ranked_gwa_ids": [
    "evaluating_compliance",
    "updating_using_knowledge",
    "interpreting_information",
    "providing_consultation",
    "communicating_internally"
  ],
  "rationale": "The interaction centers on applying a specific legal rule (federal-question jurisdiction) to the facts — checking compliance with the doctrinal requirements is the dominant AI move, supported by recalling the relevant rule (updating knowledge) and parsing what the fact pattern actually says (interpreting information). The user's questions and the AI's explanation form a consultative back-and-forth."
Example:

Benchmark Item:
Three bells ring at intervals of 36 seconds, 40 seconds and 48 seconds, \
respectively. They start ringing together at a particular time. When will \
they ring together again?
  A) After 6 minutes
  B) After 12 minutes
  C) After 18 minutes
  D) After 24 minutes
  E) none

JSON output:
{{
  "sub_steps": [
    "recognize that all three bells ring together again at the least common multiple (LCM) of their intervals",
    "factorize each interval: 36 = 2^2 * 3^2, 40 = 2^3 * 5, 48 = 2^4 * 3",
    "compute the LCM by taking the maximum power of each prime: 2^4 * 3^2 * 5 = 16 * 9 * 5 = 720 seconds",
    "convert 720 seconds to minutes: 720 / 60 = 12 minutes",
    "match against the answer choices: 12 minutes corresponds to option B",
    "select option B"
  ],
  "ranked_gwa_ids": [
    "making_decisions_solving_problems",
    "processing_information",
    "analyzing_data",
    "scheduling_work",
    "estimating_quantifiable_characteristics"
  ],
  "rationale": "The decisive sub-task is computing the LCM of three intervals (processing/analyzing the prime factorizations) and selecting the correct multiple (problem-solving). The problem is fundamentally about scheduling repeating events, which makes that GWA salient even though the format is arithmetic. Estimating quantitative characteristics supports the conversion from seconds to minutes."
Example:

Benchmark Item:
Self-efficacy (the belief that one has control over one's situation) as it \
related to job satisfaction was studied. When a group of teachers rated their \
ability to control their situation and their satisfaction with their job, the \
two variables had a correlation of 0.30. Which statement follows from this \
correlation?
  A) If you want teachers to be happy with their job, give them more control over their situation.
  B) If you want teachers to take more control over their situation, make them happier at their jobs.
  C) These two variables show a moderate negative correlation.
  D) Higher self-efficacy and higher job satisfaction tend to occur together, but neither causes the other.

JSON output:
{{
  "sub_steps": [
    "parse the setup: self-efficacy and job satisfaction were measured in a group of teachers, with r = 0.30 between them",
    "recall the meaning of the Pearson correlation coefficient: 0.30 is a positive linear association of moderate-to-weak strength, not a causal claim",
    "evaluate option A (efficacy implies satisfaction causal claim): unsupported by correlation alone, eliminate",
    "evaluate option B (satisfaction implies efficacy causal claim): also unsupported, eliminate",
    "evaluate option C (negative correlation): factually wrong because 0.30 is positive, eliminate",
    "evaluate option D (co-occurrence without causation): correctly restates what r = 0.30 implies",
    "select option D"
  ],
  "ranked_gwa_ids": [
    "analyzing_data",
    "making_decisions_solving_problems",
    "interpreting_information",
    "processing_information",
    "updating_using_knowledge"
  ],
  "rationale": "The decisive sub-task is correctly interpreting what a correlation coefficient does and does not imply. Eliminating each answer choice against that interpretation is a problem-solving step over the option set, and recalling the standard 'correlation does not imply causation' rule grounds the analysis in statistical knowledge."
}}
\end{Verbatim}
\end{tcolorbox}
\begin{tcolorbox}[enhanced,breakable,title=\textbf{Augmentation-focused Prompt}]
\begin{Verbatim}
An AI system is deployed alongside a human worker to jointly tackle this \
task — the human remains involved throughout, ranging from light oversight \
to close collaboration. Which work activities does the AI most need to be \
capable of in this collaborative setting?

Work in two steps inside the JSON output below:

STEP 1 — In `sub_steps`, imagine interacting with a user on this item and \
walk through the exchange concretely: what the user would ask, what the AI \
would do, how they go back and forth.

STEP 2 — In `ranked_gwa_ids`, rank up to 5 GWAs from the list below that \
correlate with the AI being good at this kind of interaction, ordered from \
most to least central.

Benchmark Item:
{task_text}

---

GWA List:
{gwa_list}

---

Example:

Benchmark Item:
An entrepreneur from State A decided to sell hot sauce to the public, \
labeling it "Best Hot Sauce." A company incorporated in State B and \
headquartered in State C sued the entrepreneur in federal court in State C. \
The complaint sought $50,000 in damages and alleged that the entrepreneur's \
use of the name "Best Hot Sauce" infringed the company's federal trademark. \
The entrepreneur filed an answer denying the allegations, and the parties \
began discovery. Six months later, the entrepreneur moved to dismiss for \
lack of subject-matter jurisdiction. Should the court grant the \
entrepreneur's motion?
  A) No, because the company's claim arises under federal law.
  B) No, because the entrepreneur waived the right to challenge subject-matter jurisdiction by not raising the issue initially by motion or in the answer.
  C) Yes, because although the claim arises under federal law, the amount in controversy is not satisfied.
  D) Yes, because although there is diversity, the amount in controversy is not satisfied.

JSON output:
{{
  "sub_steps": [
    "user pastes the fact pattern and four answer choices, asking which is correct",
    "AI identifies that this is a federal subject-matter jurisdiction question and surfaces the two grounds (federal question vs. diversity)",
    "AI flags the salient facts: federal trademark claim, $50,000 amount, parties from different states",
    "AI checks the live legal rule — federal-question jurisdiction has no amount-in-controversy threshold; SMJ can be raised at any time",
    "user asks whether the timing of the motion matters; AI confirms it does not for SMJ and cites the relevant procedural rule",
    "AI recommends option A and explains why federal-question grounds jurisdiction directly"
  ],
  "ranked_gwa_ids": [
    "evaluating_compliance",
    "updating_using_knowledge",
    "interpreting_information",
    "providing_consultation",
    "communicating_internally"
  ],
  "rationale": "The interaction centers on applying a specific legal rule (federal-question jurisdiction) to the facts — checking compliance with the doctrinal requirements is the dominant AI move, supported by recalling the relevant rule (updating knowledge) and parsing what the fact pattern actually says (interpreting information). The user's questions and the AI's explanation form a consultative back-and-forth."

Example:

Benchmark Item:
Self-efficacy (the belief that one has control over one's situation) as it \
related to job satisfaction was studied. When a group of teachers rated their \
ability to control their situation and their satisfaction with their job, the \
two variables had a correlation of 0.30. Which statement follows from this \
correlation?
  A) If you want teachers to be happy with their job, give them more control over their situation.
  B) If you want teachers to take more control over their situation, make them happier at their jobs.
  C) These two variables show a moderate negative correlation.
  D) Higher self-efficacy and higher job satisfaction tend to occur together, but neither causes the other.

JSON output:
{{
  "sub_steps": [
    "user shares the question and asks the AI to help reason through which option is correct",
    "AI clarifies what r = 0.30 means: a modest positive linear association, not strong, and far from causation",
    "AI walks the user through why a correlation does not establish a causal direction in either direction",
    "AI helps the user evaluate options A and B as causal claims (both unsupported by correlation alone) and option C as factually wrong (0.30 is positive, not negative)",
    "user asks 'so what is left?' — AI confirms option D restates the correlation without overclaiming causation",
    "AI explains the underlying statistical principle (correlation does not imply causation) so the user retains the reasoning, not just the answer"
  ],
  "ranked_gwa_ids": [
    "interpreting_information",
    "providing_consultation",
    "communicating_internally",
    "analyzing_data",
    "updating_using_knowledge"
  ],
  "rationale": "The interaction is fundamentally consultative: the AI walks the user through the meaning of a correlation coefficient and the limits of causal inference. Interpreting the statistical concept and communicating it back to the user dominates the AI's contribution; analytic and knowledge-recall steps support that consultation."

Example:

Benchmark Item:
Fertilizer from an agricultural area runs off into a river. The river carries \
the nutrients from this fertilizer and deposits them into an ocean bay. After \
the nutrients enter the bay, scientists monitoring the water would most \
likely see a decrease in which of these dissolved gases?
  A) oxygen
  B) nitrogen
  C) carbon dioxide
  D) carbon monoxide

JSON output:
{{
  "sub_steps": [
    "user pastes the scenario and answer choices and asks the AI to explain the underlying mechanism and pick an answer",
    "AI walks the user through the eutrophication chain: nutrient runoff causes an algal bloom, the bloom dies, microbial decomposition consumes oxygen, and hypoxia results",
    "AI rules out the other gases for the user: nitrogen is the input not the output, carbon dioxide is produced by decomposition (would increase, not decrease), carbon monoxide is not a typical aquatic gas",
    "user asks why this matters in practice — AI explains the broader environmental impact (dead zones, fish kills) so the answer is grounded in real-world consequences",
    "AI confirms option A (oxygen) and offers to elaborate on related concepts such as algal blooms or water-quality monitoring if the user wants to learn more"
  ],
  "ranked_gwa_ids": [
    "providing_consultation",
    "communicating_internally",
    "interpreting_information",
    "updating_using_knowledge",
    "evaluating_compliance"
  ],
  "rationale": "The interaction is the AI explaining a multi-step environmental mechanism to the user — consultative teaching plus communication of the reasoning chain. Recalling the eutrophication process is the knowledge backbone; helping the user evaluate which dissolved gas is affected is interpretive."
}}
\end{Verbatim}
\end{tcolorbox}

The cost of labelling one item with the (Claude) LLM judge is approximately $\$0.024$ USD.

\section{Robustness of the Welfare Pipeline}
\label{app:welfare-robustness}

Our welfare ranking depends on several pipeline choices: which LLM serves as the GWA-loading judge, how 
many GWAs are ranked per item ($k=5$ in the headline), how we 
weight GWAs within the top-$k$ ($\lambda_k=\frac{1}{k}$ in the headline), and how we aggregate GWA scores from on task-level preferences. We probe sensitivity to these choices in three ways. First, we replicate Finding~1 on a human-labeled subset of 28 OLMES items, using two independent human annotators. Second, we replicate Finding~2 on the same subset under each rater's labels. Third, we sweep over four pipeline variants---truncation depth, within-$k$ weighting, judge identity, per-GWA scoring---and report Spearman rank correlation of the resulting welfare rankings against the headline ranking.

\paragraph{Replication of Finding 1 on the human-ranked subset.} On the 28-item human-ranked subset, we recompute the per-item quadrant assignment under both human annotators and the LLM judge. \Cref{tab:human-quadrants} reports the share of loading mass in each WORKBank quadrant under each ranker-framing combination.

The qualitative pattern---most items load on the red-light or automate-only quadrants, with green-light loadings substantially smaller---holds across all three rankers and both framings. Quadrant mass differs across rankers: under augmentation, the LLM's green-light share lies between the two humans' (19.4\,\% vs.\ 13.4\,\% and 21.3\,\%), while under automation it is lower than either (4.4\,\% vs.\ 8.5\,\% and 10.2\,\%). The differences sit within the inter-human envelope on most cells.

\begin{table}[h]
\centering
\footnotesize
\caption{Quadrant share (\unit{\%}) of OLMES items under three rankers (A1, A2, three-shot-prompted Claude Opus 4.5, C), on the 28-item human-ranked subset, under both automation and augmentation framings.}
\label{tab:human-quadrants}
\begin{tabular}{@{}ll S S S S@{}}
\toprule
{Framing} & {Ranker} & {{Green}} & {{Auto-only}} & {{Augment-only}} & {{Red}} \\
\midrule
\multirow{3}{*}{Automation}   & A1     &  8.5 & 49.1 & 21.7 & 20.8 \\
                              & A2     & 10.2 & 51.0 & 23.5 & 15.3 \\
                              & C &  4.4 & 41.6 & 20.4 & 33.6 \\
\addlinespace
\multirow{3}{*}{Augmentation} & A1     & 13.4 & 32.3 & 24.4 & 29.9 \\
                              & A2     & 21.3 & 25.4 & 23.0 & 30.3 \\
                              & C & 19.4 & 26.1 & 26.1 & 28.4 \\
\bottomrule
\end{tabular}
\end{table}

\paragraph{Replication of Finding 2 on the human-ranked subset.} Under all 
three rankers and both framings, the finding of math items being Pareto inferior to general-knowledge items on our panel
replicates on the human-ranked subset, see \Cref{tab:human-pareto}. The welfare 
Pareto frontier on this subset contains zero math items across all 
ranker-framing combinations, and all math items are 
dominated by at least one general-knowledge item. The mean welfare gap 
between general-knowledge and math items is positive on both axes for all 
rankers, ranging from \num{0.56} to \num{+2.23} on $w^\mathrm{auto}$ and from \num{+0.09} to \num{+1.37} on $w^\mathrm{aug}$.

\begin{table}[h]
\centering
\footnotesize
\caption{Replication on the human-ranked subset across three rankers. $\Delta w^\mathrm{auto}$ and $\Delta w^\mathrm{aug}$ report the mean welfare gap between general-knowledge and math items. \enquote{Frontier (math/total)} counts math items on the Pareto frontier out of total frontier items on the subset. \enquote{\% math dominated} is the fraction of math items Pareto-dominated by at least one general-knowledge item.}
\label{tab:human-pareto}
\sisetup{table-format=+1.2, table-number-alignment=center, detect-weight=true}
\begin{tabular}{@{}l S S c S[table-format=3.0]@{}}
\toprule
{Ranker} & {$\Delta w^{\mathrm{auto}}$} & {$\Delta w^{\mathrm{aug}}$} & {Math on frontier} & {{\% math dominated}} \\
\midrule
A1       & +2.23 & +0.09 & 0 / 2 & 100 \\
A2       & +1.83 & +0.80 & 0 / 3 & 100 \\
C   & +0.56 & +1.37 & 0 / 3 & 100 \\
\bottomrule
\end{tabular}
\end{table}

\paragraph{Sensitivity to pipeline choices.} We sweep over four variants of the welfare aggregation pipeline: (i) varying the top-$k$ truncation depth ($k=3$, $k=4$, $k=5$ with reciprocal-rank weighting); (ii) replacing reciprocal-rank weighting with uniform within-$k$ weighting (at $k=5$); (iii) replacing threshold-based per-GWA scoring ($\mathbb{P}[A_w \geq 3.5]$) with mean-based scoring ($\bar{A}_w$); and (iv) substituting the LLM judge from Claude Opus 4.5 to GPT-5.2 (at $k=5$ with reciprocal-rank weighting). Spearman rank correlation of each variant's welfare ranking against the headline ranking is reported in \Cref{tab:welfare_robustness}.

Two patterns hold. First, the ranking is stable to truncation depth: $\rho = 0.91$ at $k=3$ and $\rho = 0.95$ at $k=4$ on $w^\mathrm{auto}$, with similar stability on $w^\mathrm{aug}$. The headline finding does not depend on the specific choice of $k$. In addition, switching from threshold-based to mean-based per-GWA scoring retains a strong but not perfect correlation ($\rho = 0.71$ on $w^\mathrm{auto}$, $\rho = 0.90$ on $w^\mathrm{aug}$): the threshold preserves discriminative signal that the central-limit averaging in $\bar{A}_w$ partly compresses. Second, the ranking is sensitive to two choices we view as substantive rather than incidental: (i) uniform within-$k$ weighting ($\rho = 0.33$ on $w^\mathrm{auto}$) discards the rank information that reciprocal weighting treats as load-bearing under automation, while retaining it under augmentation ($\rho = 0.88$ on $w^\mathrm{aug}$). This indicates that the rank-1 GWA carries more  differentiating signal under automation than under augmentation: $w^\mathrm{auto}$ depends substantially on rank-1, while $w^\mathrm{aug}$ is 
distributed more evenly across the top-5. (ii) Switching to a different LLM judge ($\rho = 0.67$ on $w^\mathrm{auto}$, $\rho = 0.50$ on $w^\mathrm{aug}$) introduces the same annotator-level variation documented in the human-judge calibration (\Cref{app:prompts}). Within the family of choices that share our load-bearing assumptions (rank-inverse weighting, single judge), the welfare ranking is stable.

\begin{table}[t]
\centering
\small
\caption{Robustness of per-item welfare scores to the ranking aggregation rule and to the LLM judge. Each row reports Spearman rank correlations between $w^{\mathrm{auto}}_i$ (resp.\ $w^{\mathrm{aug}}_i$) under the variant specification and under the baseline ($k=5$, reciprocal weights $\lambda_k = 1/k$, Claude Opus 4.5). Top four rows perturb the aggregation rule; bottom row swaps the LLM judge.}
\label{tab:welfare_robustness}
\sisetup{table-format=1.2, table-number-alignment=center}
\begin{tabular}{@{}l S S@{}}
\toprule
{Variant} & {$\rho(w^{\mathrm{auto}})$} & {$\rho(w^{\mathrm{aug}})$} \\
\midrule
\multicolumn{3}{@{}l}{\emph{Aggregation rule (Claude Opus 4.5 judge)}} \\
\quad $k=3$, reciprocal weights      & 0.91 & 0.87 \\
\quad $k=4$, reciprocal weights      & 0.95 & 0.89 \\
\quad $k=5$, uniform weights         & 0.33 & 0.88 \\
\quad mean-based per-GWA agg.        & 0.71 & 0.90 \\
\addlinespace
\multicolumn{3}{@{}l}{\emph{LLM judge ($k=5$, reciprocal weights)}} \\
\quad GPT-5.2                        & 0.67 & 0.50 \\
\bottomrule
\end{tabular}
\end{table}

\section{Base Rate's Influence on Variance}\label{app:additional_noise_results}

\Cref{tab:noise-quant-4B} showed that raw seed-to-seed noise $\hat\sigma_i$ is concentrated on HellaSwag, PIQA, and ARC-Challenge, and is smallest on MMLU, CommonsenseQA, and MathQA. We now ask whether this ranking reflects a genuine difference in benchmark stability, or a mechanical artifact of where the model's accuracy sits.

\paragraph{Setup.} For each item $i$ we observe a binary correctness score $y_{ims} \in \{0,1\}$ across $S=8$ seeds and $K=10$ checkpoints. Write $\hat p_i$ for the pooled mean and $\hat\sigma_i$ for the pooled across-seed standard deviation, both as defined in \Cref{sec:setup-noise}. Items are scored by acc norm, the standard length-normalized scoring rule for multiple-choice items: the model's prediction is the candidate $c$ that maximizes the per-character log-probability $\overline{\log p}(c) \coloneqq \log p(c \mid \text{prompt}) / \mathrm{len}(c)$, and $y_{ims} = 1$ iff this argmax is the gold answer.

\paragraph{Two candidate mechanisms.} A binary score has variance $\hat p(1-\hat p)$ that peaks at $\hat p = 0.5$ and shrinks toward the tails. So benchmarks on which the model is near chance ($\hat p \approx 0.25$ for 4-way multiple choice) will mechanically look quieter than benchmarks on which the model is near $\hat p = 0.5$, regardless of any genuine stability difference. We strip this out by normalizing:
\[
Z_i \coloneqq \frac{\hat\sigma_i}{\sqrt{\hat p_i (1 - \hat p_i)}}.
\]
Pure binomial sampling at $K = 10$ would give $Z_i \approx 1/\sqrt{K} \approx 0.32$; values above this indicate excess seed-to-seed noise beyond the sampling floor.

A second mechanism operates on items the model gets right. The gap between the gold answer's score and the best distractor's score is
\[
m_i \;\coloneqq\; \mathrm{median}\,
   \frac{\overline{\log p}_{\text{gold}}}
        {\overline{\log p}_{\text{runner-up}}}
\]
measured in nats per character. Items with small $m_i$ sit near the acc norm decision boundary, where small seed-induced shifts in continuation likelihoods can flip the argmax.

\begin{table}[t]
\caption{Decomposition of per-family seed-to-seed noise on the PolyPythia-410M panel ($S=8$ seeds, $K=10$ checkpoints). \emph{Bernoulli component}: median $\hat p_i$, median $\hat\sigma_i$, and the Bernoulli-normalized $Z_i = \hat\sigma_i / \sqrt{\hat p_i(1-\hat p_i)}$. The binomial floor at $K=10$ is $Z \approx 0.32$. \emph{Margin component}: on items the model gets right ($n_{\mathrm{won}}$), median margin $m_i$ in nats per character, and within-family Pearson correlation $r$ between $m_i$ and $Z_i$.}
\label{tab:noise-decomp}
\centering
\small
\sisetup{table-format=1.3, table-number-alignment=center}
\begin{tabular}{@{}l S[table-format=5.0] S S S S[table-format=4.0] S S[table-format=-1.2]@{}}
\toprule
 & & \multicolumn{3}{c}{Bernoulli component} & \multicolumn{3}{c}{Margin component} \\
\cmidrule(lr){3-5}\cmidrule(lr){6-8}
\textbf{Family} & {$n$} & {$\hat p$} & {$\hat\sigma$} & {$Z$} & {$n_{\mathrm{won}}$} & {$m$ (nats/char)} & {$r(m, Z)$} \\
\midrule
\textsc{HellaSwag} & 2029  & 0.512 & 0.261 & 0.615 & 1034 & 0.003 & -0.08 \\
\textsc{PIQA}      & 837   & 0.592 & 0.208 & 0.475 &  494 & 0.031 &  0.01 \\
\textsc{ARC-C}     & 549   & 0.310 & 0.179 & 0.421 &  140 & 0.012 & -0.18 \\
\textsc{MathQA}    & 1889  & 0.235 & 0.092 & 0.219 &  257 & 0.097 & -0.05 \\
\textsc{MMLU}      & 13732 & 0.161 & 0.071 & 0.198 & 2566 & 0.404 & -0.57 \\
\textsc{CSQA}      & 963   & 0.154 & 0.070 & 0.195 &  110 & 0.331 & -0.41 \\
\bottomrule
\end{tabular}
\end{table}

\paragraph{Bernoulli component.} \Cref{tab:noise-decomp} shows that part of the raw $\hat\sigma$ ranking is mean-driven. PolyPythia-410M is near chance on MMLU, CSQA, and MathQA (median $\hat p \in [0.15, 0.24]$) but close to $\hat p = 0.5$ on HellaSwag and PIQA, which mechanically inflates $\hat\sigma$ on the latter two. Normalization collapses MMLU, CSQA, and MathQA to $Z_i \approx 0.20$, \emph{below} the binomial floor of $0.32$, so their seed-to-seed answers are essentially deterministic at this scale. HellaSwag, by contrast, sits at $Z \approx 0.62$---about twice the floor---so its excess noise survives the mean correction.

\paragraph{Margin component.} The remaining excess noise on HellaSwag, PIQA, and ARC-Challenge tracks $m_i$. HellaSwag's median margin is $0.003$ nats per character; MMLU's is $0.404$, two orders of magnitude larger. The within-family correlation is more telling than the medians. On MMLU, $m_i$ correlates with $Z_i$ at $r = -0.57$: tight-margin items are reliably the noisy ones, as expected if margin sets the per-item noise floor. On HellaSwag this correlation collapses to $r = -0.08$, consistent with adversarial filtering having pushed \emph{all} items to a tight-margin ceiling, so margin no longer separates noisy from quiet items.

The residual seed-to-seed noise on HellaSwag, PIQA, and ARC-Challenge therefore traces to benchmark construction rather than to model capability or item-format artifacts: distractors are by design semantically close to gold in log-probability space, so small seed-dependent perturbations flip the acc norm argmax.

\include{checklist}

\end{document}

%% file: checklist.tex
\section*{NeurIPS Paper Checklist}
\begin{enumerate}

\item {\bf Claims}
    \item[] Question: Do the main claims made in the abstract and introduction accurately reflect the paper's contributions and scope?
    \item[] Answer: \answerYes{}
    \item[] Justification: The abstract and introduction state two contributions: (i) a closed-form characterization of welfare-optimal benchmark weights as a function of three primitives ($w$, $M$, $\Sigma$), proved in \Cref{sec:model} and \Cref{thm:optimal_v}; and (ii) an audit framework that ranks items along each of these three axes, instantiated on OLMES items in \Cref{sec:empirical} and \Cref{sec:results}. Limitations of the empirical pipeline---in particular the diagonal-$C$ assumption, the cross-panel scope, and the per-axis (rather than canonical-$v^\star$) headline---are surfaced in \Cref{sec:conclusion}.

\item {\bf Limitations}
    \item[] Question: Does the paper discuss the limitations of the work performed by the authors?
    \item[] Answer: \answerYes{}
    \item[] Justification: \Cref{sec:conclusion} contains a dedicated limitations discussion covering the diagonal-$C$ assumption (substitutes/complements channel theoretical rather than measured), the normative status of $w$ (pro-worker stance is a position, not a fact), the cross-panel scale mismatch between the cost panel (EvoLM 4B) and the noise panel (PolyPythias 410M), and the deliberate choice to deliver per-axis rankings rather than a single canonical aggregation.

\item {\bf Theory assumptions and proofs}
    \item[] Question: For each theoretical result, does the paper provide the full set of assumptions and a complete (and correct) proof?
    \item[] Answer: \answerYes{}
    \item[] Justification: The closed-form $v^\star = (M+r\Sigma)^{-1}Mw$ in \Cref{thm:optimal_v} is derived in \Cref{sec:model} from explicitly stated assumptions ($M \succeq 0$, $\Sigma \succ 0$, CARA preferences, Gaussian noise, linear contract). The Gaussianity and linear-quadratic assumptions are relaxed in \Cref{app:foa}, where \Cref{app:foa} states the score-concavity condition under which the first-order approach is valid and provides a full proof.

    \item {\bf Experimental result reproducibility}
    \item[] Question: Does the paper fully disclose all the information needed to reproduce the main experimental results of the paper to the extent that it affects the main claims and/or conclusions of the paper (regardless of whether the code and data are provided or not)?
    \item[] Answer: \answerYes{}
    \item[] Justification: \Cref{sec:empirical} specifies the data sources (WORKBank, EvoLM 4B suite, PolyPythias 410M panel, OLMES), all eligibility filters (anchor-probability and slope thresholds, mean-bound and degeneracy filters), the per-item OLS specification, the pooled-variance estimator, and the bootstrap protocol with replicate counts. \Cref{app:costs} fully derives the dollar-effort unit prices $\kappa_k$. \Cref{app:prompts} gives the verbatim LLM-judge prompts.

\item {\bf Open access to data and code}
    \item[] Question: Does the paper provide open access to the data and code, with sufficient instructions to faithfully reproduce the main experimental results, as described in supplemental material?
    \item[] Answer: \answerYes{}
    \item[] Justification: All inputs are publicly available (WORKBank, EvoLM, PolyPythias, OLMES, O*NET, BLS), and we link to our replication package in the abstract.

\item {\bf Experimental setting/details}
    \item[] Question: Does the paper specify all the training and test details (e.g., data splits, hyperparameters, how they were chosen, type of optimizer) necessary to understand the results?
    \item[] Answer: \answerYes{}
    \item[] Justification: This is an audit study rather than a training study; the relevant ``hyperparameters'' are the eligibility thresholds ($\hat p \in [0.05, 0.95]$ and non-negative slope for cost, $\bar p \in [0.10, 0.90]$ and $\hat\sigma^2 \ge 10^{-6}$ for noise), the welfare threshold ($A_w \ge 3.5$), the LLM-judge models (GPT-5.2 and Claude Opus 4.5) and methods, and the bootstrap replicate count ($B{=}1000$). All are specified in \Cref{sec:empirical}.

\item {\bf Experiment statistical significance}
    \item[] Question: Does the paper report error bars suitably and correctly defined or other appropriate information about the statistical significance of the experiments?
    \item[] Answer: \answerYes{}
    \item[] Justification: Both ranking tables (\Cref{tab:cost-quant-4B}, \Cref{tab:noise-quant-4B}) report $[2.5\%, 97.5\%]$ bootstrap rank confidence intervals over $B{=}1000$ replicates. For cost, replicates resample residuals from the per-item OLS; for noise, replicates resample seeds with replacement within each checkpoint cluster.

\item {\bf Experiments compute resources}
    \item[] Question: For each experiment, does the paper provide sufficient information on the computer resources (type of compute workers, memory, time of execution) needed to reproduce the experiments?
    \item[] Answer: \answerYes{}
    \item[] Justification: Compute usage is dominated by inference passes over OLMES items on the EvoLM 4B grid and PolyPythias 410M panel (existing public checkpoints) and by LLM-judge API calls for GWA loadings. Aggregate inference and API costs are not reported in the present version; we plan to add them in a revision.
    
\item {\bf Code of ethics}
    \item[] Question: Does the research conducted in the paper conform, in every respect, with the NeurIPS Code of Ethics \url{https://neurips.cc/public/EthicsGuidelines}?
    \item[] Answer: \answerYes{}
    \item[] Justification: The work uses only existing public datasets (worker preferences elicited and released by \citet{shao2025futureworkaiagents}, public model panels, and standard ML benchmark items) and does not involve new human-subjects data collection. The pro-worker normative stance is surfaced explicitly in \Cref{sec:setup-welfare} as a position rather than a fact.

\item {\bf Broader impacts}
    \item[] Question: Does the paper discuss both potential positive societal impacts and negative societal impacts of the work performed?
    \item[] Answer: \answerYes{}
    \item[] Justification: The paper's main object of study---how benchmark design shapes lab incentives---is itself a societal-impact analysis: \Cref{sec:related-work} and \Cref{sec:model} discuss how misaligned aggregation distorts effort allocation away from welfare-relevant capabilities (Goodharting). Potential negative impacts include reifying a particular normative stance (pro-worker welfare) as the basis for benchmark design, which we surface in the limitations of \Cref{sec:conclusion}. Potential positive impacts include providing benchmark designers, auditors, and regulators with quantitative per-item annotations they can use to diagnose welfare-misalignment in existing benchmarks.
    
\item {\bf Safeguards}
    \item[] Question: Does the paper describe safeguards that have been put in place for responsible release of data or models that have a high risk for misuse (e.g., pre-trained language models, image generators, or scraped datasets)?
    \item[] Answer: \answerNA{}
    \item[] Justification: The paper releases neither pre-trained models nor scraped datasets. The planned artifacts are per-item welfare/cost/noise rankings of public benchmark items and the analysis code that produced them, neither of which carries dual-use or misuse risk beyond the underlying public datasets.

\item {\bf Licenses for existing assets}
    \item[] Question: Are the creators or original owners of assets (e.g., code, data, models), used in the paper, properly credited and are the license and terms of use explicitly mentioned and properly respected?
    \item[] Answer: \answerYes{}{}
    \item[] Justification: All datasets and model panels used are cited at point of use: WORKBank \citep{shao2025futureworkaiagents}, EvoLM \citep{qi2026evolm}, PolyPythias \citep{wal2025polypythias}, OLMES \citep{gu2025olmes}, and O*NET \cite{onet2026}. Each is used within the terms under which it was publicly released. Specific license names are enumerated in the replication package at \url{https://github.com/stair-lab/principal-agent-benchmarks}.

\item {\bf New assets}
    \item[] Question: Are new assets introduced in the paper well documented and is the documentation provided alongside the assets?
    \item[] Answer:  \answerYes{}
    \item[] Justification: Our code is linked as an anonymous repository in the abstract.

\item {\bf Crowdsourcing and research with human subjects}
    \item[] Question: For crowdsourcing experiments and research with human subjects, does the paper include the full text of instructions given to participants and screenshots, if applicable, as well as details about compensation (if any)? 
    \item[] Answer: \answerNA{}
    \item[] Justification: No new human-subjects data is collected. The worker-preference data we use is taken from the publicly released WORKBank dataset \citep{shao2025futureworkaiagents}, which reports its own elicitation protocol and compensation. The GWA-ranking calibration in \Cref{app:prompts} is a methodological calibration on existing public benchmark items, not crowdsourced data collection.

\item {\bf Institutional review board (IRB) approvals or equivalent for research with human subjects}
    \item[] Question: Does the paper describe potential risks incurred by study participants, whether such risks were disclosed to the subjects, and whether Institutional Review Board (IRB) approvals (or an equivalent approval/review based on the requirements of your country or institution) were obtained?
    \item[] Answer: \answerNA{}
    \item[] Justification: As above, no new human-subjects data is collected.

\item {\bf Declaration of LLM usage}
    \item[] Question: Does the paper describe the usage of LLMs if it is an important, original, or non-standard component of the core methods in this research? Note that if the LLM is used only for writing, editing, or formatting purposes and does \emph{not} impact the core methodology, scientific rigor, or originality of the research, declaration is not required.
    \item[] Answer: \answerYes{}
    \item[] Justification: \Cref{app:prompts} detail the LLM-judge protocol used to elicit per-item GWA loadings, including the two judges (GPT-5.2 and Claude Opus 4.5), the two methods (direct top-$k$ ranking and $k{=}10$ sparse pairwise with Borda aggregation), and the two framings (automation under H1 and augmentation under H2--H4). The verbatim prompts appear in \Cref{app:prompts}, and calibration against a human-expert reference set is reported in \Cref{tab:gwa_ranking_auto_aug}.

\end{enumerate}